\pdfoutput=1

\documentclass[11pt]{article}

\usepackage[final]{acl}

\usepackage{times}
\usepackage{latexsym}

\usepackage[T1]{fontenc}

\usepackage[utf8]{inputenc}

\usepackage{microtype}

\usepackage{inconsolata}

\usepackage{graphicx}

\usepackage{booktabs}

\usepackage{todonotes}

\usepackage{multirow}

\usepackage{enumitem}
\usepackage{svg}
\usepackage{graphicx}
\usepackage{booktabs}

\usepackage{float}

\usepackage[table]{xcolor}

\definecolor{myblue}{RGB}{194,233,255}

\definecolor{giallotenue}{rgb}{1.0, 1.0, 0.8}

\usepackage{amssymb}

\usepackage{colortbl}
\usepackage{pgfmath}
\usepackage[most]{tcolorbox}

\usepackage{url}

\usepackage[most]{tcolorbox} 
\tcbset{
  myboxstyle/.style={
    colback=gray!10!white,   
    colframe=gray!70!black,  
    fonttitle=\bfseries,     
    coltitle=white,          
    boxrule=0.3mm,           
    arc=2mm,                 
    left=1mm, right=1mm,     
    top=1mm, bottom=1mm      
  }
}

\usepackage[table]{xcolor}   
\usepackage{colortbl}        
\usepackage{booktabs}        
\usepackage{siunitx}         
\usepackage{pgf}             
\usepackage{ifthen}          

\usepackage{multirow}
\usepackage{booktabs}
\usepackage{makecell}
\usepackage{subcaption}
\usepackage{todonotes}

\newcommand{\perc}[2]{%
  \pgfmathsetmacro{\increase}{((#2 - #1) / #1 * 100)}%
  \pgfmathparse{round(\increase)}%
  \pgfmathtruncatemacro{\increaseInt}{\pgfmathresult}%
  \pgfmathtruncatemacro{\displayPerc}{abs(\increaseInt)}%
  \pgfmathsetmacro{\colorlevel}{max(0, min(100, \displayPerc * 3))}%
  \ifdim \increase pt > 0pt
    \def\arrowSymbol{$\uparrow$}%
    \def\cellColor{green}%
  \else
    \def\arrowSymbol{$\downarrow$}%
    \def\cellColor{red}%
  \fi
  \edef\temp{\noexpand\cellcolor{\cellColor!\colorlevel} \num[round-precision=0]{#2}\, \arrowSymbol\displayPerc\%\ }%
  \temp%
}

\newcommand{\percsecnew}[2]{%
  \pgfmathsetmacro{\increase}{((#2 - #1) / #1 * 100)}%
  \pgfmathparse{round(\increase)}%
  \pgfmathtruncatemacro{\increaseInt}{\pgfmathresult}%
  \pgfmathtruncatemacro{\displayPerc}{abs(\increaseInt)}%
  \pgfmathsetmacro{\colorlevel}{max(0, min(100, \displayPerc * 3))}%
  \ifdim \increase pt > 0pt
    \def\arrowSymbol{$\uparrow$}%
    \def\cellColor{green}%
  \else
    \def\arrowSymbol{$\downarrow$}%
    \def\cellColor{red}%
  \fi
  \edef\temp{\noexpand\cellcolor{\cellColor!\colorlevel} \arrowSymbol\displayPerc\%\, \num[round-precision=0]{#2}  }%
  \temp%
}

\newcommand{\percnew}[2]{%
  \pgfmathsetmacro{\increase}{((#2 - #1) / #1 * 100)}%
  \pgfmathtruncatemacro{\increaseInt}{\increase}%
  \pgfmathtruncatemacro{\displayPerc}{abs(\increaseInt)}%
  \ifdim \increase pt > 0pt
  \pgfmathsetmacro{\colorlevel}{max(0, min(100, \displayPerc * 3))}%
    \def\arrowSymbol{$\uparrow$}%
    \def\cellColor{green}%
  \else
    \pgfmathsetmacro{\colorlevel}{max(0, min(100, \displayPerc * 1))}%
    \def\arrowSymbol{$\downarrow$}%
    \def\cellColor{red}%
  \fi
  \edef\temp{\noexpand\cellcolor{\cellColor!\colorlevel} \num[round-precision=0]{#2}\,\arrowSymbol }%
  \temp%
}

\definecolor{cadmiumgreen}{rgb}{0.0, 0.42, 0.24}

\title{Detecting Winning Arguments \\ with Large Language Models and Persuasion Strategies}

\author{
 \textbf{Tiziano Labruna\textsuperscript{1}},
 \textbf{Arkadiusz Modzelewski\textsuperscript{1,2}},
 \textbf{Giorgio Satta\textsuperscript{1}},
 \textbf{Giovanni Da San Martino\textsuperscript{1}}
\\
\\
 \textsuperscript{1}University of Padua,
 \textsuperscript{2}Polish-Japanese Academy of Information Technology
\\
}

\begin{document}
\maketitle
\begin{abstract}
Detecting persuasion in argumentative text is a challenging task with important implications for understanding human communication. 
This work investigates the role of persuasion strategies---such as \textit{Attack on reputation, Distraction}, and \textit{Manipulative wording}---in determining the persuasiveness of a text. 
We conduct experiments on three annotated argument datasets: \textit{Winning Arguments} (built from the \textit{Change My View} subreddit), \textit{Anthropic/Persuasion}, and \textit{Persuasion for Good}. 
Our approach leverages large language models (LLMs) with a Multi-Strategy Persuasion Scoring approach that guides reasoning over six persuasion strategies.
Results show that strategy-guided reasoning improves the prediction of persuasiveness. 
To better understand the influence of content, we organize the \textit{Winning Argument} dataset into broad discussion topics and analyze performance across them. We publicly release this topic-annotated version of the dataset to facilitate future research. Overall, our methodology demonstrates the value of structured, strategy-aware prompting for enhancing interpretability and robustness in argument quality assessment.

\end{abstract}

\begin{figure*}[t]
    \centering
    \includegraphics[width=0.95\textwidth]{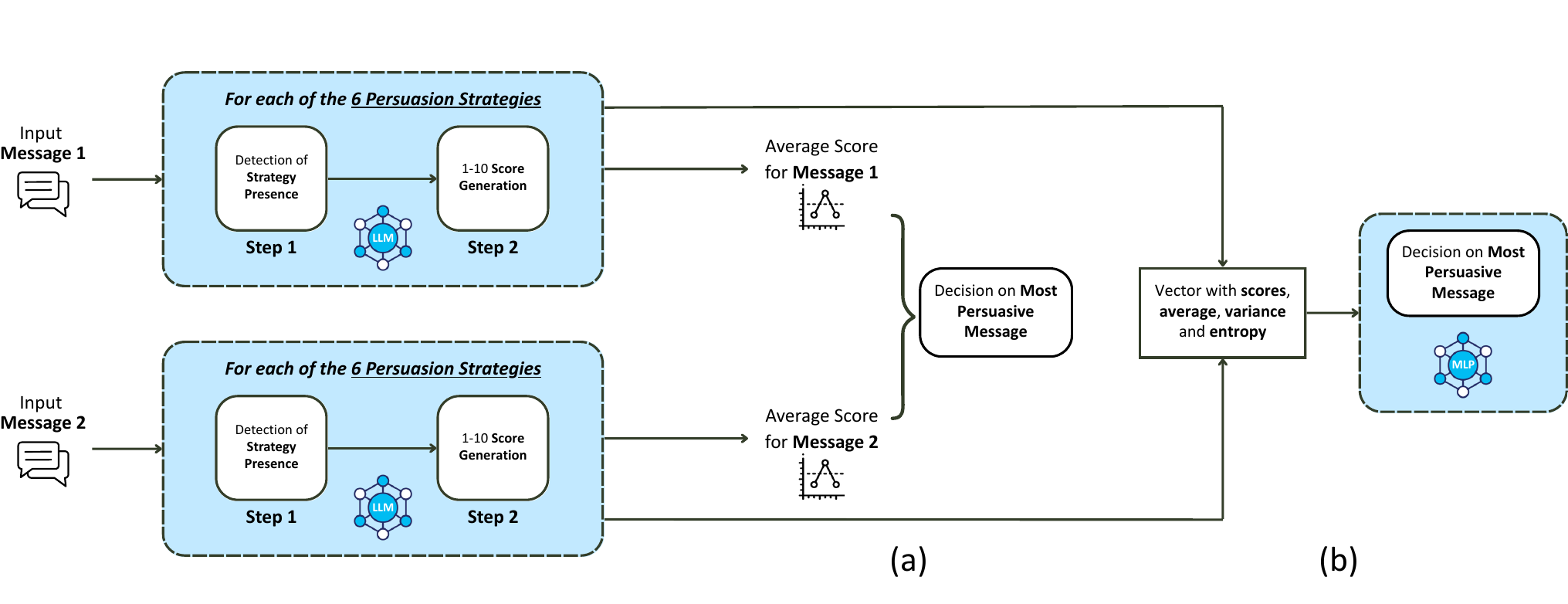}
    \caption{Overview of the MS-PS framework. Each of the two input messages is independently analyzed by a language model across six persuasion strategies. For each strategy, the model first generates an explanation assessing the presence of the strategy, followed by a 1–10 persuasiveness score. In the \textbf{MS-PS-AVG} variant (a), the more persuasive message is identified as the one with the higher average score. In the \textbf{MS-PS-MLP} variant (b), each message is represented by a feature vector consisting of the six individual scores plus their average, variance, and entropy, which is fed to a trained MLP classifier to predict which message is more persuasive.}
    \label{fig:persuasive_flow}
\end{figure*}

\section{Introduction}

In an era where online discussions shape public opinion, understanding what makes certain arguments more persuasive and ultimately successful is becoming increasingly important. The significance of this problem has been widely recognized, inspiring extensive research across multiple disciplines \cite{saenger2024autopersuade, hoeken2012arguing, maio2014social, van2015speech}. Identifying arguments that effectively change opinions is particularly relevant for political debate analysis, decision-making support, and applications in computational social science \cite{popkin1991reasoning, ziems2024can}.

Previous research on predicting winning arguments---i.e., those that successfully convince a reader to adopt a certain opinion---has mainly focused on assessing argument quality, evaluating convincingness through annotated argument pairs, and analyzing interaction dynamics \cite{habernal2016makes, gleize2019you, tan2016winning}. Recent studies have also examined how large language models (LLMs) detect persuasive arguments \cite{rescala2024can, ziems2024can}. To our knowledge, no prior work has explored the role of specific persuasion strategies in identifying winning arguments. In this study, we address this gap by investigating whether particular persuasion strategies serve as discriminative signals for winning arguments, moving beyond general linguistic or interaction patterns.

To investigate the usefulness of persuasion strategies in winning arguments detection, we use the \textit{Winning Arguments} (\textsc{WA}) dataset \cite{tan+etal:16a}, collected from the \textit{Change My View} subreddit, an online forum on Reddit where users propose discussion topics and present and evaluate arguments.
To enable more granular, topic-aware analysis, we further introduce the \textit{\textbf{T}opics \textbf{W}inning \textbf{A}rguments} (\textsc{TWA}) dataset, a topic-annotated extension of \textsc{WA}. 
\textsc{TWA} organizes discussions by topic, providing a finer-grained examination of argument persuasiveness.


Motivated by the Persuasion-Augmented Chain of Thought (PCoT) method for strategy-aware reasoning in disinformation detection \cite{modzelewski2025pcot}, we propose \textbf{M}ulti-\textbf{S}trategy \textbf{P}ersuasion \textbf{S}coring (\textbf{MS-PS}), adopting a well-established taxonomy of six persuasion strategies, previously used in NLP research \cite{dimitrov2024semeval, piskorski2023news, piskorski2023semeval}.
MS-PS leverages LLMs to perform structured, strategy-specific reasoning: for each persuasion strategy, the model first generates a textual analysis and then assigns a numerical persuasiveness score, resulting in one score per strategy for each message. These scores are used in two ways: (i) averaging across strategies to select the message with the higher mean, and (ii) training a multilayer perceptron that predicts the more persuasive message from the LLM-generated scores (see Figure \ref{fig:persuasive_flow}). While averaging the scores provides a simple system to assess how the use of persuasion strategies correlates with persuasiveness, the multilayer perceptron aims to capture more nuanced interplay between multiple strategies and their impact on persuasiveness.
Unlike PCoT, which injects all strategies jointly and focuses on their detection, our approach fully disentangles strategies by using independent prompts for each one, produces continuous strategy-specific scores, and learns a non-linear aggregation of these signals. These design choices reduce cross-strategy interference and yield a more interpretable and expressive representation of persuasion.
We validate MS-PS both against human annotations and on  
two additional datasets: \textit{Anthropic/Persuasion} \cite{durmus2024persuasion} and \textit{Persuasion for Good} \cite{wang2019persuasion}, demonstrating its effectiveness across diverse domains. 

Our main contributions are: 
(I) we are the first to exploit information about persuasion strategies in a learning model with the goal of detecting winning arguments; 
(II) we introduce \textsc{MS-PS}, a zero-shot framework producing strategy-specific persuasiveness scores usable both directly and as features for supervised models; (III) the release of the \textsc{TWA} dataset, a topic-annotated extension of Winning Arguments enabling fine-grained, topic-aware analysis; (IV) extensive validation of \textsc{MS-PS} across datasets, showing robustness, effectiveness on contamination-free data, and scalability beyond binary tasks.

\section{Related Work}

Early work on persuasion detection used lexical, syntactic, and sentiment features combined with rule-based or classical ML techniques~\cite{tan2016winning, hidey2017analyzing, lukin2017argument}. These methods offered useful signals but lacked robustness and struggled to capture discourse-level strategies.

The advent of LLMs enabled richer, context-aware representations and led to strong gains via fine-tuning and in-context learning for persuasion tasks~\cite{bassi2024, li2024uncovering, nayak2024analyzing}. LLMs have been used to produce interpretable features~\cite{li2024lmeme}, to support multi-label and multilingual persuasion classification~\cite{purificato2023apatt, hromadka2023kinitveraai, roll2024greybox}, and to model belief change~\cite{Hoang2025}.

Prior works have also applied LLMs to detecting manipulative content such as propaganda~\cite{sprenkamp2023large, hasanain2024can}; these studies show models like GPT-4 can match state-of-the-art systems but still face limits in fine-grained reasoning and cross-language generalization. A growing line of research focuses specifically on identifying winning arguments, with LLMs performing competitively with humans~\cite{rescala2024can}, and extensions to multilingual/domain-specific settings using reasoning-based prompting~\cite{jose2025llms}.

We build on these trends by introducing a structured approach: rather than treating persuasiveness as a monolithic property, we perform structured reasoning prompting through LLMs to analyze specific rhetorical strategies and produce interpretable per-strategy scores that support both zero-shot and supervised prediction.

\section{Datasets and Tasks}
\label{sec:datasets_tasks}

In this work, we primarily focus on the \textit{Winning Arguments} (WA) dataset (Section \ref{sec:wa_dataset}), a well-established benchmark built from natural online debates, which provides a controlled binary setup for studying persuasive success.  
To enable richer analysis, we extend this dataset with our topic-annotated version, \textit{Topics Winning Arguments} (TWA) (Section \ref{sec:twa_dataset}), which allows investigation of how persuasiveness varies across different domains of discussion.  
Finally, to ensure that our findings generalize beyond WA and are not affected by potential data contamination, we also evaluate on two additional datasets released more recently: \textit{Anthropic/Persuasion} (Section \ref{sec:anthropic_dataset}) and \textit{Persuasion for Good} (Section \ref{sec:pfg_dataset}).  

\subsection{Winning Arguments}
\label{sec:wa_dataset}

The \textit{Winning Arguments} (\emph{WA}) dataset \cite{tan+etal:16a} was originally constructed from the \textit{Change My View} subreddit,\footnote{https://reddit.com/r/changemyview} an online forum where users post opinions and invite others to challenge them. A counterargument is marked as successful when the original poster explicitly acknowledges being persuaded, signaled by the awarding of a $\Delta$.

Each data instance consists of a pair of messages taken from the same discussion: one persuasive (delta-awarded) and one non-persuasive. The task is to predict which message is the \textit{winning} one, i.e., the one that successfully changed the original poster’s view. 

To construct balanced pairs, each persuasive argument (i.e., a rooted path-unit that received a $\Delta$) was matched with an unsuccessful one from the same thread, chosen for high topical similarity. This similarity was measured using Jaccard overlap between the non-stopword sets of the initial replies, ensuring that paired arguments discuss the same subject matter. 

The final dataset contains 4,263 pairs, divided into 2,746 for training, 710 for validation, and 807 for testing. Its design makes it a challenging benchmark, as persuasiveness is shaped not only by the content of arguments but also by how effectively they resonate with the original poster’s perspective.

\subsection{The Topics Winning Arguments Dataset}
\label{sec:twa_dataset}

To enable topic-aware analysis of persuasive arguments, we introduce the \textbf{Topics Winning Arguments (TWA)} dataset, derived from the \textit{Winning Arguments} corpus. TWA retains all samples and splits from WA, with the exception of one pair\footnote{Pair p\_1601} removed from the validation set due to a data error where the persuasive and non-persuasive messages were identical.

To organize the data into coherent topical domains, we applied BERTopic \cite{grootendorst2022bertopic}, a transformer-based topic modeling framework that clusters arguments using contextual embeddings. This procedure yielded four high-level topics:  
(1)~\textbf{Food and Culture},  
(2)~\textbf{Religion and Ethical Debates},  
(3)~\textbf{Economics and Politics},  
(4)~\textbf{Gender, Sexuality, and Minority Rights}.  

Further implementation details on the topic modeling procedure are provided in Appendix~\ref{appendix:twa_topic_modeling}, along with illustrative argument examples (Table~\ref{tab:twa_examples}). Additional statistics on message length and lexical diversity per topic are reported in Appendix~\ref{appendix:twa_stats}.

We release TWA to encourage future work on domain-aware argument mining and to support robust generalization across diverse discussion themes.

\subsection{Anthropic/Persuasion}
\label{sec:anthropic_dataset}

The \textbf{Anthropic/Persuasion} (or just \emph{Anthropic}) dataset \cite{durmus2024persuasion} was introduced to study the persuasiveness of human- and model-generated arguments. Each sample consists of a triple: an original claim, a candidate argument (produced either by a human or an LLM), and a reader’s self-reported agreement ratings (on a 1-7 Likert scale) with the claim before and after reading the argument. The associated task is to predict the final agreement rating, given the claim, the argument, and the initial rating. The dataset contains 3,939 samples in total and was originally released as a single split. For our experiments, we partitioned it into training (2,757), development (591), and test (591) sets.

\subsection{Persuasion for Good}
\label{sec:pfg_dataset}

The \textbf{Persuasion for Good} (\emph{P4G}) dataset \cite{wang2019persuasion} consists of dialogues (collected via Amazon Mechanical Turk) where a persuader attempts to convince a persuadee to donate money to charity. Persuasiveness is measured directly by the actual donation amount, providing a concrete behavioral outcome rather than a subjective judgment. The task associated with this dataset is to predict the donated amount, given the persuader’s turns in the dialogue. The dataset contains 1,017 conversations and was originally released as a single split. For our experiments, we partitioned it into training (711), development (153), and test (153) sets.

\section{Methodology}
\label{sec:methodology}

We introduce \textbf{MS-PS} (\textbf{M}ulti-\textbf{S}trategy \textbf{P}ersuasion \textbf{S}coring), a structured framework designed to evaluate and compare the persuasiveness of messages by explicitly modeling and scoring rhetorical strategies. Unlike simple direct comparison methods, MS-PS decouples reasoning and scoring, encouraging more consistent and interpretable judgments. 
This section focuses on the task formulation for the WA dataset. However, as we will show in Section~\ref{subsec:extension_new_domains}, the adaptation to the \textit{Anthropic} and \textit{P4G} datasets is trivial.
Figure~\ref{fig:persuasive_flow} illustrates the full pipeline: given a pair of messages responding to the same post, each message is processed independently through a two-step prompting protocol across six persuasive strategies.


\begin{figure}[tp!]
  \fbox{
    \begin{minipage}{0.95\columnwidth}
      \scriptsize 
      \colorbox{myblue}{\textbf{Attack on reputation [AR]}} - the argument does not address the topic itself but targets the participant (personality, experience, etc.) to question and/or undermine their credibility. The object of the argumentation can also refer to a group of individuals, an organization, an object, or an activity.\\
      
      \colorbox{myblue}{\textbf{Justification [J]}} - the argument is made of two parts, a statement and an explanation or appeal, where the latter is used to justify and/or to support the statement.\\
      
      \colorbox{myblue}{\textbf{Simplification [S]}} - the argument excessively simplifies a problem, usually regarding the cause, the consequence, or the existence of choices.\\
      
      \colorbox{myblue}{\textbf{Distraction [D]}} - the argument takes focus away from the main topic or argument to distract the reader.\\
      
      \colorbox{myblue}{\textbf{Call [C]}} - the text is not an argument, but an encouragement to act or to think in a particular way.\\
      
      \colorbox{myblue}{\textbf{Manipulative wording [MW]}} - the text is not an argument per se, but uses specific language, which contains words or phrases that are either non-neutral, confusing, exaggerating, loaded, etc., in order to impact the reader emotionally.
    \end{minipage}
  }
  \caption{Description of the six persuasion strategies used in our experiments.}
  \label{fig:persuasion_strategies_tax}
\end{figure}

\subsection{Step 1: Strategy-Aware Reasoning} 

We adopt a well-established and widely used taxonomy of six persuasion strategies \cite{dimitrov2024semeval, piskorski2023news, piskorski2023semeval}: \emph{Attack on Reputation}, \emph{Distraction}, \emph{Manipulative Wording}, \emph{Simplification}, \emph{Justification}, and \emph{Call} (the last directly encourages the reader to take a position or action, rather than constructing an argument). These strategies are illustrated in more detail in Figure~\ref{fig:persuasion_strategies_tax}.

For each strategy, we prompt an LLM to generate a natural language analysis of the message’s rhetorical structure, explicitly focusing on whether the strategy is present and how it contributes to the message’s persuasiveness. The prompts encourage careful reasoning and instruct the model to identify a strategy only when there is clear textual evidence.

This method builds on the Persuasion-Augmented Chain of Thought (PCoT) framework \cite{modzelewski2025pcot}, which introduced strategy-guided prompting for detecting persuasion techniques.

Full prompt details are provided in Appendix~\ref{appendix:strategy_analysis_prompt}. Following the standard structure used by large language models, the prompt consists of a \textit{system message} that defines the general context and analysis strategy and a \textit{user message} that provides the specific directions on the task, including the post’s title and body, the candidate message, and the instructions for critical analysis.

\subsection{Step 2: Strategy Scoring from Reasoning}

After generating the reasoning for each strategy, we prompt the model to assign a persuasiveness score from 1 to 10, grounded in the explanation produced in the previous step. The choice of a 1--10 scale, rather than alternative ranges, is validated through dedicated experiments, which we discuss in detail in Appendix~\ref{appendix:persuasion_scale_choice}. The prompt includes the original post, the candidate message, and the model’s reasoning, and requests a numerical score accompanied by a brief justification.
Full prompt templates are provided in Appendix~\ref{appendix:strategy_scoring_prompt}.
Consequently, each message receives six individual persuasion scores, one for each strategy.

\subsection{MS-PS-AVG: Zero-Shot Aggregation}
\label{sec:avg-method}

In the \textbf{MS-PS-AVG} variant, we average the six strategy scores to obtain an overall persuasiveness estimate for each message. The message with the higher average is predicted as the more persuasive one. In the rare case of a tie (same score average for both messages),  we apply \textit{message perturbation}: we rephrase the messages using LLMs and recompute their scores until a preference emerges. Rewriting is attempted iteratively, starting with light lexical variation and increasing the intensity of rewriting if necessary. Prompts range from minor rephrasing to complete stylistic neutralization. All rewriting prompt templates and implementation details are included in Appendix \ref{appendix:tie_breaking}. 

\subsection{MS-PS-MLP: Learning a Persuasion Function}
\label{sec:mlp-method}

While averaging provides a simple aggregation heuristic, it assumes that all strategies contribute equally to persuasiveness and that higher scores are always better. In practice, however, the presence of a particular strategy may be more or less critical depending on the context.

To model more nuanced patterns, we design \textbf{MS-PS-MLP}, as a learned classifier that predicts which of two messages is more persuasive based on their strategy scores.

For each message, we construct a 9-dimensional feature vector comprising the six individual strategy scores, along with the \emph{average}, \emph{variance}, and \emph{entropy} of the scores for each message. More information on why these 3 additional features where selected and how they are computed, is presented in Appendix \ref{appendix:MLP-features}.


We then train a supervised model (a multilayer perceptron, or \emph{MLP}) over the LLM-generated persuasion scores. 
For \textsc{WA}, the model is framed as a binary classifier: it takes as input the concatenation of the strategy-specific scores from the two messages (18 features) and outputs which of the two was the persuasive one. 
For \emph{Anthropic} and \emph{P4G}, the task is formulated as regression: the MLP takes as input the 9-dimensional vector of scores for a single message and predicts a continuous persuasiveness value. 
This learned approach allows the model to capture non-linear interactions and strategy-specific weightings that simple heuristics, such as averaging, cannot encode.


%

\section{Experimental Setup}
\label{sec:experimental_setup}

This section outlines the experimental setup used to evaluate the effectiveness of our proposed framework, \textbf{MS-PS}, described in Section~\ref{sec:methodology}. We compare MS-PS against a set of baseline approaches designed to assess message persuasiveness, starting from simpler formulations and gradually increasing the complexity and interpretability of the evaluation.

\subsection{Language Models}
\label{subsec:LLMs}

We evaluated five LLMs spanning both open-source and proprietary APIs: \textit{Gemma-3-12B} and \textit{Llama-3.1-8B} (open-source), as well as \textit{Gemini-1.5-Flash-02}, \textit{Gemini-2.0-Flash}, and \textit{OpenAI-o3} (accessed via API). All models were evaluated in a zero-shot setting without additional fine-tuning or supervision. To ensure comparability, we standardized preprocessing of inputs, enforced consistent prompting formats, and adopted similar decoding strategies across models. 
The open-source models were executed locally on high-memory GPUs, while proprietary models were accessed via Google and OpenAI’s APIs. General implementation details are described in Appendix~\ref{appendix:llm_impl}, whereas Appendix~\ref{appendix:llms} provides technical information specific to the APIs used.

\subsection{Direct Comparison}

We initially tested a direct comparison format in which the model is prompted with both candidate messages and asked to select the more persuasive one (see Appendix \ref{appendix:direct_comparison_prompt} for the prompt template). While this setup is intuitive and mirrors human evaluation tasks, we observed a strong and systematic \textit{positional bias}: models tended to favor the second message regardless of content (see Appendix \ref{appendix:direct_comparison_results}).

This phenomenon has been previously documented in the literature \cite{shi2024judging}, where LLMs exhibit a preference for one specific options in comparison tasks. This undermined the reliability of this format, making it inappropriate for our evaluation goals.

We also experimented with a perturbation-based variant of this setting, inspired by \citet{ziems2024can}, where each message was paraphrased multiple times before comparison. As discussed in Appendix \ref{app:perturbation_experiment}, this alternative did not yield better results, with model performance remaining close to random.

\subsection{Independent Scoring Baselines}
\label{subsec:independent_scores}

To avoid the positional bias observed in direct comparison, we transitioned to single-message prompts in which each candidate is evaluated independently. Specifically, each message is passed to the LLM with a prompt asking for a persuasiveness score on a 1–10 scale. The scores for the two messages are then compared to determine the more persuasive one. In the case of a tie, we apply the same rephrasing-based resolution strategy used in MS-PS (see Appendix \ref{appendix:tie_breaking}).

We define four variants of this scoring approach, each progressively enriching the model’s input:

\begin{itemize}
\item \textbf{Independent Scoring:} The message is presented alone, and the model is instructed to return a persuasiveness score from 1 to 10, based solely on the message content.

\item \textbf{+ Context:} The model is given the title and body of the original post in addition to the message. It is asked to rate the persuasiveness of the message based on this context, returning only a numerical score from 1 to 10.

\item \textbf{+ Explanation:} The message is shown in isolation, and the model must both provide a 1–10 persuasiveness score and briefly justify its choice with a natural language explanation.

\item \textbf{+ Context + Explanation:} The title and body of the original post are provided along with the message. The model returns a 1–10 score and an explanation of its reasoning.
\end{itemize}

These single-prompt setups serve as stronger baselines than direct comparison, and help assess how context and reasoning affect model judgments. The specific prompts used in each variant are reported in Appendix~\ref{appendix:independent_scoring_prompts}.

\subsection{Experiments on Winning Arguments}
\label{subsec:msps_eval}

We evaluated the two variants of our proposed \textbf{MS-PS} framework (AVG and MLP) on all five LLMs in a zero-shot setting. We experimented with multiple prompt formulations to guide strategy-aware reasoning and selected the prompt yielding the highest accuracy on the validation set for each model. To assess the consistency  of strategy scoring across models, we computed inter-model agreement scores using Cohen’s kappa; the results show moderate agreement between most model pairs (see Appendix \ref{app:agreement} for details).

In addition, to verify that the observed performance gains are not merely driven by prompt complexity, we compare \textsc{MS-PS} against two generic baselines (a simple persuasiveness analysis and an explicit chain-of-thought variant) and further control for input length (Appendices~\ref{appendix:strategy_role},~\ref{appendix:token_length}). The results consistently show that structured guidance based on persuasion strategies yields superior performance across models. We further analyze how individual persuasion strategies contribute differently across topics and models, providing a more fine-grained view of strategy effectiveness (Appendix~\ref{appendix:strategy_impact}). 

For the MLP variant, we first computed MS-PS strategy scores on the training, development, and test splits for each LLM. Then, for each model, we trained a separate binary classifier using a three-layer neural network that takes as input an 18-dimensional vector per message pair (comprised of \textit{mean}, \textit{variance}, and \textit{entropy} of strategy scores) and predicts the more persuasive message.

We performed an extensive grid search over architectural and training hyper-parameters, including hidden layer sizes, learning rates, regularization factors, and early stopping configurations. Each model was trained for up to 300 epochs with early stopping based on dev set performance. The full search space and best hyper-parameters per model are reported in Appendix~\ref{appendix:msps_grid}.

\begin{table}[t]
\centering
\scriptsize
\renewcommand{\arraystretch}{1}
\setlength{\tabcolsep}{3pt}
\begin{tabular}{lccccc}
    \toprule
    \multicolumn{1}{c}{\textbf{Strategy}} & \multicolumn{5}{c}{\textbf{Model Accuracy}} \\
             & LLama & Gemma & Gemini-1.5 & Gemini-2 & o3 \\
             \midrule
             Independent Scoring & 56.66 & 53.78 & 56.01 & 56.13 & 55.51 \\
             + Context & 59.48 & 59.48 & 61.96 & 60.84 & 58.98\\
            + Explanation & 56.26 & 60.72 & 56.38 & 59.11 & 58.24 \\
            + Context + Explan. & 54.52 & 58.24 & 61.09 & 61.46 & 60.35 \\
            \hline \\
            MS-PS-AVG & 60.72 & 62.83 & 60.72 & 61.83 & 60.59 \\
        MS-PS-MLP & \textbf{61.34} & \textbf{63.69} & \textbf{63.07} & \textbf{62.70} & \textbf{64.53} \\
        \bottomrule
\end{tabular}
\caption{Accuracy (\%) of different prompting strategies and models on the Winning Arguments test set. "Llama" is model Llama-3.1-8B, "Gemma" is Gemma-3-12B, "o3" is OpenAI-o3. Best results per model in bold.}
\label{tab:persuasiveness-results}
\end{table}

\subsubsection{Results and Discussion}

Table \ref{tab:persuasiveness-results} reports the accuracy of the four baseline settings along with the two proposed approaches (MS-PS-AVG and MS-PS-MLP) evaluated on the test-set of the Winning Arguments dataset. All methods are implemented using the five LLMs detailed in Section \ref{subsec:LLMs}, and evaluated on the task of identifying the more persuasive message in each pair. 
Baseline results prove that when evaluating messages in isolation, all models perform only marginally above random guessing (50\%). This is expected: lacking access to the original post or reasoning, models have limited basis for assessing persuasiveness. 
Adding the original post (title + body) consistently improves performance, highlighting the importance of conversational context in persuasive judgments. Asking the model to provide a justification for the given score (+ Explanation) yields mixed results: while Gemma-3-12B benefits significantly (+6.94\%), the gains are smaller or absent for the other models. Surprisingly, combining both context and explanation does not consistently outperform the use of context alone. For instance, Gemini-1.5 performs slightly worse with both additions (61.09\%) compared to context alone (61.96\%). This drop could be due to longer inputs or reasoning inconsistencies introduced in the explanation step. These results highlight that simply increasing input richness is not guaranteed to improve model judgment.

Both MS-PS variants outperform all baselines across all models, with the exception of Gemini-1.5, which already performs strongly with the + Context baseline. 
These results validate the effectiveness of our strategy-aware scoring approach and confirm that structured reasoning based on rhetorical strategies enhances persuasiveness evaluation.

The \texttt{MLP} variant consistently outperforms the \texttt{AVG} version across all evaluated models. However, this performance gain is generally not statistically significant (McNemar's test: $p>0.05$) for the majority of the models compared.
An exception is observed for the \texttt{OpenAI-o3} model. This model initially recorded the lowest performance in the \texttt{AVG} configuration, but it became the overall best-performing model in the \texttt{MLP} variant. This specific improvement is confirmed to be statistically significant ($p=0.03$).
This indicates that simply averaging strategy scores is not sufficient to fully capture persuasiveness. Instead, the neural network layer in the MLP variant can learn to recognize patterns in the combination and interplay of different rhetorical strategies, allowing it to better judge which message is more persuasive.
For additional insight into how models distribute their strategy scores across persuasive and non-persuasive messages, we refer the reader to Appendix \ref{appendix:strategy-distributions}, which visualizes these distributions for all six MS-PS strategies.

\subsection{Validating MS-PS Scores on Human Annotations}

To assess the reliability of our two-step scoring process, we conducted an experiment aimed at validating the MS-PS strategy scores against human annotated persuasion labels.
We used a dataset from Task 3 of SemEval 2023 \cite{piskorski2023semeval}, which contains 536 English news articles. Each article is annotated for the presence or absence of one or more of the six persuasion strategies described in Figure \ref{fig:persuasion_strategies_tax}, which are the same used for MS-PS. We compared two approaches:
\begin{enumerate}
    \item \textbf{Single-Prompt Classification}: For each strategy, we asked an LLM to directly predict whether the strategy was present (yes/no) in a given article using a short, targeted prompt (the full prompt is included in Appendix~\ref{appendix:single_prompt_validating}).
    \item \textbf{MS-PS Scoring}: Following the methodology of our proposed approach, presented in Section~\ref{sec:methodology}, for  each strategy, we first injected the model with knowledge about the strategy and asked it to generate an analytical paragraph evaluating the presence of the strategy. Based on this analysis, the model then assigned a score from 1 to 10 indicating the strength of the strategy in the article (the full prompt is included in Appendix~\ref{appendix:msps_validating}). This yielded a six-dimensional vector of strategy scores per article.
\end{enumerate}

To convert the continuous scores into binary predictions, we selected the optimal threshold for each strategy based on validation performance (the thresholds used are presented in Table~\ref{tab:msps_thresholds}). 

\begin{table}[ht]
    \centering
    \renewcommand{\arraystretch}{1.2}
\setlength{\tabcolsep}{5pt} 
    \scriptsize 
    \begin{tabular}{l c}
        \toprule
         Method & F\textsubscript{1} Micro \\
        \midrule
        MS-PS &  \percsecnew{0.664}{0.722}$\pm0.035$ \\
        Single-Prompt & 0.664$\pm0.030$ \\ 
                \bottomrule
    \end{tabular}
    \caption{Micro-averaged F\textsubscript{1} scores with standard deviation across four LLMs on the SemEval 2023 dataset, comparing MS-PS and single-prompt classification.}
    \label{tab:persuasion_step_binary_vs_multi}
\end{table}

Results (summarized in Table~\ref{tab:persuasion_step_binary_vs_multi} and extensively presented in Appendix~\ref{appendix:strategy_validation}) show that MS-PS systematically outperforms the single-prompt classification approach across all six strategies, indicating that the reasoning step on the persuasion strategies leads to more accurate detection. These findings support the validity of our scoring methodology and confirm that the MS-PS-generated scores reflect meaningful assessments of persuasive content.

\subsection{Experiments on Anthropic and Persuasion for Good}
\label{subsec:extension_new_domains}

To further validate the robustness of our method, we extended MS-PS-MLP (our best performing variant) to the \textit{\textbf{Anthropic}} and \textit{\textbf{Persuasion for Good}} datasets. Unlike the Winning Arguments dataset, these tasks are formulated as regression problems: predicting a reader’s post-argument rating shift (\emph{Anthropic}) or the amount of money donated after a persuasion dialogue (\emph{P4G}). 

For these experiments, we slightly modified our neural network architecture to process a single input representation (six strategy scores plus aggregated statistics) and output a single continuous value. This modification highlights the scalability of MS-PS, as the same strategy-aware framework can generalize to tasks with varying prediction targets.
We compared MS-PS-MLP against two single-prompt baselines: a basic prompt directly asking to predict the outcome (Baseline-1), and a refined prompt adjusted to improve the simpler prompt (Baseline-2). Across both datasets, MS-PS-MLP consistently outperformed these baselines, demonstrating that the benefits of strategy-aware reasoning extend beyond the CMV setting. Importantly, in the case of the Anthropic dataset, which was released after the cutoff dates of our models, these results also confirm that our improvements are not due to data contamination. 

Table~\ref{tab:msps_regression_mse} summarizes the improvements in root mean squared error (RMSE) over the baselines. Full details of these experiments are reported in Appendix \ref{appendix:extension_new_domains}.

\begin{table}[ht]
    \centering
    \renewcommand{\arraystretch}{1.2}
    \setlength{\tabcolsep}{6pt}
    \scriptsize
    \begin{tabular}{lcc}
        \toprule
         \textbf{Method} & \textbf{Anthropic} & \textbf{Persuasion for Good} \\
        \midrule
        Baseline-1 & 1.55 & 152.15 \\
        Baseline-2 & 1.44 & 102.07 \\
        \textbf{MS-PS-MLP} & \textbf{0.82} & \textbf{4.63} \\
        \bottomrule
    \end{tabular}
    \caption{RMSE comparison (the lower the better) across methods on Anthropic and Persuasion for Good datasets. Baselines are averaged on the results of Gemma-3-12B and OpenAI-o3.}
    \label{tab:msps_regression_mse}
\end{table}

\subsection{Topic-Based Evaluation}
\label{sec:topic_eval}

Building on the topical annotations provided in the TWA dataset (see Section~\ref{sec:twa_dataset}), we investigate whether persuasion behaves differently across discussion domains, a question that cannot be addressed using the original WA dataset due to the absence of topic labels. Specifically, we analyze how the performance of the MS-PS framework (measured as accuracy in identifying the delta-awarded message) varies across the four thematic domains defined in TWA, comparing both the AVG and MLP variants.

\begin{table}[h]
\centering
\scriptsize
\setlength{\tabcolsep}{4pt}
\renewcommand{\arraystretch}{1.2}
\begin{tabular}{lcccc}
\toprule
\textbf{Model} & \shortstack{\textbf{Food} \\ \textbf{\& Culture}} 
               & \shortstack{\textbf{Religion} \\ \textbf{\& Ethics}} 
               & \shortstack{\textbf{Economics} \\ \textbf{\& Politics}} 
               & \shortstack{\textbf{Gender, Sex.} \\ \textbf{\& Minorities}} \\
\midrule
\multicolumn{5}{c}{\textbf{MS-PS-AVG}} \\
\midrule
LLaMA-3.1-8B    & 64.46\% & 64.37\% & 58.97\% & 57.08\% \\
Gemma-3-12B     & \textbf{69.88\%} & 62.64\% & 59.40\% & 61.37\% \\
Gemini-1.5      & 61.45\% & 61.49\% & 60.68\% & 59.66\% \\
Gemini-2        & 66.87\% & 64.94\% & 59.40\% & 58.37\% \\
OpenAI-o3       & 61.45\% & 58.05\% & 61.11\% & 61.37\% \\
\midrule
\multicolumn{5}{c}{\textbf{MS-PS-MLP}} \\
\midrule
LLaMA-3.1-8B    & 65.06\% & 65.52\% & 59.40\% & 57.51\% \\
Gemma-3-12B     & 68.67\% & \textbf{67.82\%} & 59.83\% & 60.94\% \\
Gemini-1.5      & 63.86\% & 66.67\% & 60.68\% & 62.23\% \\
Gemini-2        & 66.27\% & 66.67\% & 58.97\% & 60.94\% \\
OpenAI-o3       & 64.46\% & 66.09\% & \textbf{64.96\%} & \textbf{64.38\%} \\
\bottomrule
\end{tabular}
\caption{Accuracy of MS-PS-AVG and MS-PS-MLP across the four debate topics. Bold indicates the highest accuracy per topic.}
\label{tab:topic_generalization}
\end{table}

Table~\ref{tab:topic_generalization} reports results on the topic-specific subsets of the TWA test set, which includes 166 pairs for \emph{Food and Culture}, 174 for \emph{Religion and Ethical Debates}, 234 for \emph{Economics and Politics}, and 233 for \emph{Gender, Sexuality, and Minority Rights}. Across models, higher accuracy is generally observed for the first two topics, while performance drops for the latter two. This suggests that persuasion strategies are more effective in domains grounded in everyday experience or moral reasoning, whereas their impact is reduced in more polarized or sensitive domains, where opinions tend to be more entrenched.

More broadly, these results confirm that persuasion effectiveness varies systematically across topics, motivating the need for topic-aware analysis. Leveraging the same topic labels, we further analyze how the relative importance of individual strategies changes across domains (Appendix~\ref{appendix:strategy_impact}), an effect that is only observable thanks to the TWA dataset and that naturally opens the door to future work on topic-conditioned strategy weighting.

\section{Conclusions and Future Works}

In this work, we addressed the task of detecting persuasive messages from text, focusing primarily on the challenging setup provided by the \textit{Winning Arguments} dataset. 
Our experiments demonstrated that prompting LLMs to analyze specific persuasion strategies and using their outputs as structured features for a downstream classifier leads to significant improvements over simpler baselines. 
We also contributed a new version of the dataset annotated with topic categories, which we release publicly to support future research on topic-aware persuasion detection. The analysis on the different topics provided further insights into how the nature of the discussion impacts model performance, highlighting areas where detecting persuasion is particularly difficult.
Future work could explore fine-tuning strategies, expanding the set of rhetorical techniques considered, and evaluating transferability to other argumentative settings.

\section*{Limitations}

While our study shows promising results in detecting persuasive messages using large language models and strategy-based analyses, several limitations remain.

First, our approach depends on the quality of the datasets used (\textit{Winning Arguments}, \textit{Anthropic/Persuasion}, and \textit{Persuasion for Good}). While these datasets provide a controlled evaluation setup, they may reflect the culture, writing style, and norms of a limited population (e.g., the \textit{Change My View} community). As such, generalization to other domains or conversational settings may be limited.

Second, the strategy-based scoring mechanism depends on the models' ability to accurately recognize and interpret rhetorical strategies based solely on text descriptions. Errors or inconsistencies in how models apply these criteria can introduce noise into the feature representations used for classification.

Third, while we treat the \textit{winning} message as the more persuasive one for evaluation purposes, it is important to recognize that persuasiveness is inherently subjective and can be influenced by individual factors such as prior beliefs or personal preferences. Nonetheless, the large scale and consistent structure of the datasets help mitigate this effect.

We believe that addressing these limitations in future work could further enhance the robustness and applicability of persuasion detection systems.

\section*{Ethical Considerations}

Our work uses Reddit discussions to study persuasive language, relying on large language models to generate strategy-based scores and training a lightweight classifier on these outputs. While we do not train language models directly, the LLMs used may reflect biases from their pretraining data, which can influence how persuasion strategies are detected and interpreted.

The Winning Arguments dataset includes publicly available content that may touch on sensitive or controversial topics. To protect user privacy, we report only aggregated results and never disclose usernames or direct quotes.

We will release our topic-annotated version of the dataset (TWA) under a CC BY-NC-ND 4.0 license to support non-commercial research. No crowdsourcing was involved in the creation of this resource; topic labels were generated automatically using BERTopic and refined through manual inspection by the authors.

Leveraging large language models often requires substantial computational resources, which can raise environmental concerns. However, our approach minimized computational demand by relying on inference through API-based access to LLMs, without training any large models from scratch. We trained only a lightweight multilayer perceptron classifier on the strategy scores, with each run taking approximately 30 seconds. We trained this model using a grid search described in Appendix \ref{appendix:msps_grid}. All training was run on Nvidia A40 GPUs provided by the university for research and educational use.

While our goal is to improve understanding of persuasive communication, we acknowledge the risk of misuse, such as optimizing manipulative messaging. We encourage responsible use of this work and further research into its societal implications.

\bibliography{anthology,custom}

\appendix

\begin{table*}[ht]
\centering
\renewcommand{\arraystretch}{1.3}
\setlength{\tabcolsep}{8pt}
\rowcolors{2}{gray!10}{white}
\begin{tabular}{p{3cm} p{5.5cm} p{7.5cm}}
\toprule
\rowcolor{gray!25} 
\textbf{Topic} & \textbf{Title} & \textbf{Body (excerpt)} \\
\midrule
\textbf{Food and Culture} & 
\small CMV: I think Benn Wyatt is right that calzones are superior to Pizza in almost every way. & 
\small Benn Wyatt made several good points about the virtues of calzones. For those that, god forbid, don't already know: [..]  
1) Calzones are easy to transport cleanly. [..]  
2) Calzones maintain their temperature better. [..]  
3) As any New Yorker will tell you, the proper way to eat pizza is to fold it in half [..]  
\textit{To me, the answer is clear. We should become a calzone nation.} \\
\midrule
\textbf{Religion and Ethical Debates} & 
\small CMV: Practicing absolute pacifism is immoral & 
\small Clarification of statement: I am using the definition of absolute pacifism as ``total nonviolence and unconditional rejection of all forms of warfare''.  
My opinion relies on the following ideas:  
* Sometimes a small act of violence can prevent more violence. [..]  
* An absolute pacifist relies on others to commit violence for their safety. [..]  
* Going to war can, in some cases, be more morally justifiable than refusing. \\
\midrule
\textbf{Economics and Politics} & 
\small I believe the Republican Party will die within the next 100 years and will be replaced by the Libertarian Party. CMV & 
\small It has recently become fairly clear that the Republican Party is disconnected with the younger generation. Millennials are skeptical of religion, supportive of gay rights and climate action, and view Republicans as out of touch. [..]  
I predict that within the next couple decades, Republicans will decline in support. [..]  
They will ultimately accept their fate and rebrand as libertarians. \\
\midrule
\textbf{Gender, Sexuality, and Minority Rights} & 
\small CMV: Women who are anti-feminism do not have internalized misogyny. & 
\small First let me clarify that anti-feminism isn't necessarily anti-women's rights, but against belief in patriarchy and institutional misogyny. [..]  
Feminists often try to silence criticism by saying women have internalized misogyny. I don't believe this is the case. [..]  
Saying a woman has internalized misogyny discredits her lived experience. \\
\bottomrule
\end{tabular}
\caption{Representative examples of titles and excerpts of argument bodies for each topic in the TWA dataset. Bodies are truncated for readability.}
\label{tab:twa_examples}
\end{table*}

\section{Implementation Details of TWA Topic Modeling}
\label{appendix:twa_topic_modeling}

In order to develop the TWA dataset, as a subdivision by topic of the Winning Arguments dataset, we applied BERTopic \cite{grootendorst2022bertopic}, a transformer-based topic modeling framework. Here we provide a detailed description of the preprocessing, clustering, and post-processing steps.

\subsection{Motivation for Four Topics}
The decision to divide the dataset into four clusters was informed by a preliminary analysis. Initially, we experimented with a range of $3$---$10$ clusters and observed that $4$ clusters offered a good balance between granularity and interpretability. While $3$ clusters created a too strong merge of distinct discussion themes, more clusters produced overly fine-grained topics that were less stable across different random initializations. Therefore, $4$ clusters were chosen to provide consistent and meaningful high-level topics that cover the main domains of online debates in \textit{Change My View}.

\subsection{Text Preprocessing}
Before modeling, each post’s title and body were combined into a single text string. We applied light preprocessing to remove noise and standardize the input:
\begin{itemize}[nosep]
    \item Removal of URLs, mentions, and punctuation.
    \item Conversion to lowercase.
    \item Removal of stop words from both standard English stopword lists and a custom set including tokens like \texttt{cmv} and \texttt{delta}.
    \item Removal of words shorter than three characters.
\end{itemize}
Additionally, extraneous introductory text such as ``*Hello, users of CMV!*'' in the post body was removed to avoid noise in the topic clusters.

\subsection{Topic Modeling and Balanced Clustering}
To ensure consistency and comparability across topics, we replaced the default HDBSCAN clustering in BERTopic with a custom \texttt{BalancedKMeans} implementation. This method enforces roughly equal cluster sizes by initially assigning points to clusters based on proximity to cluster centers while respecting a maximum capacity, followed by iterative refinement.  

The BERTopic vectorizer was configured as follows:
\begin{itemize}[nosep]
    \item \texttt{TfidfVectorizer} with \texttt{ngram\_range=(1,2)}, \texttt{min\_df=3}, and \texttt{max\_df=0.9}.
    \item Minimum topic size: 15 documents.
    \item Number of topics: 4 (matching the balanced KMeans clusters).
    \item Probability estimates were calculated for all documents to allow post-hoc topic balancing if needed.
\end{itemize}

\subsection{Impact of Balanced Clustering}
Enforcing balanced cluster sizes helped prevent dominant topics from overwhelming smaller but meaningful discussion themes. While this approach introduced some coarseness in the cluster assignments, we found through manual inspection that the noise introduced was minimal and did not compromise the interpretability of the resulting topics. The standard deviation of topic sizes was approximately 30--40 documents, indicating that the balance constraint effectively reduced disparities across clusters without excessive distortion.

\subsection{Topic Labeling}
After clustering, topic labels were assigned post-hoc by manually inspecting the top 20 most representative tokens for each cluster. Two experts conducted this process collaboratively, yielding the following high-level topics: (1)~Food and Culture (1113 pairs), (2)~Religion and Ethical Debates (1057 pairs), (3)~Economics and Politics (1056 pairs), (4)~Gender, Sexuality, and Minority Rights (1036 pairs). While coarse-grained, these labels capture distinct domains commonly discussed in online debates and enable topic-aware analysis of persuasive strategies. 

Table~\ref{tab:twa_examples} provides illustrative examples of titles and excerpts of argument bodies for each of the four topics, highlighting the diversity and interpretability of the clusters.

\subsection{Reproducibility}
All preprocessing, clustering, and topic assignment scripts, including the \texttt{BalancedKMeans} implementation, will be made publicly available upon paper acceptance. Researchers will be able to reproduce the topic assignments and modify the clustering constraints or preprocessing steps if desired.

\section{TWA Topic Statistics}
\label{appendix:twa_stats}

\begin{table}[H]
\centering
\small
\begin{tabular}{ccccc}
\toprule
\textbf{Topic} & \makecell{\textbf{\#}\\\textbf{Pairs}} & \makecell{\textbf{Avg.}\\\textbf{Chars}} & \makecell{\textbf{Avg.}\\\textbf{Words}} & \makecell{\textbf{Avg.}\\\textbf{Unique Words}} \\
\midrule
1 & 1113 & 1738.43 & 301.55 & 175.77 \\
2 & 1057 & 1795.10 & 307.68 & 175.90 \\
3 & 1056 & 1908.78 & 323.87 & 187.13 \\
4 & 1036 & 1812.54 & 311.14 & 178.34 \\
\bottomrule
\end{tabular}
\caption{Topic-wise statistics for TWA. Averaged values consider both successful and unsuccessful messages for all pairs. The topic numbering correspond to the subdivision detailed in Section~\ref{sec:twa_dataset}.}
\label{tab:topic_stats}
\end{table}

Table~\ref{tab:topic_stats} reports detailed statistics for each of the four TWA topics, including the number of argument pairs, average character length, word count, and lexical diversity (unique words). Despite being roughly balanced in size, the topics vary slightly in verbosity and lexical richness, with Topic~3 (Economics and Politics) showing the highest average length and vocabulary. This variability highlights the need for models to generalize across stylistic and topical differences.

\section{Strategy Analysis Prompt}
\label{appendix:strategy_analysis_prompt}

In the first step of MS-PS, the LLM is prompted to generate a natural language analysis of the presence and use of a given persuasion strategy in a message. The prompt is as follows:

\begin{quote}
\small
\textbf{System:}  
You are an assistant who detects persuasion in text. Persuasive text is characterized by a specific use of language in order to influence readers. Your expertise and focus is on one high-level persuasion approach called \texttt{strategy\_name}. This high-level persuasion approach can be described as follows: \texttt{strategy\_description}

\textbf{User:}  
Given the title and body of an original post, and a message intended to change the perspective of the original poster, critically evaluate its persuasive potential. Identify strengths and weaknesses in its argumentation, structure, and rhetorical strategies. Additionally, analyze whether the message employs the high-level persuasion approach \texttt{strategy\_name}. Explain how \texttt{strategy\_name} appears or doesn't appear in the message. Be conservative in your assessment and if you are not fully certain that the technique is used, assume it is not present.
\end{quote}

Each strategy is assessed independently using this template.

\section{Strategy Scoring Prompt}
\label{appendix:strategy_scoring_prompt}

In the second step of MS-PS, the LLM is asked to score the persuasive effectiveness of the message on a scale from 1 to 10, based on the previously generated analysis. The prompt used is:

\begin{quote}
\small
\textbf{System:}  
You are a Persuasion Detector, your goal is to detect the degree of persuasiveness of a message ranging from 1 to 10, where persuasion is the potential of changing someone's opinion. You will be prompted with the title and the body of the original poster and the message that tries to make the original poster change their view, as well as an analysis on the persuasion of the message.

\textbf{User:}  
Given the title and the body of an original poster and a message that tries to make the original poster change their view, as well as an analysis on persuasion strategies used in the message, you have to respond with a number from 1 to 10 based on the degree of persuasiveness of the message, preceded by a brief explanation on why you gave that score. Give your answer in the form of a dictionary:\\
\texttt{\{"explanation": "Your answer. Brief explanation on the reasoning that you have followed.", "response": "Your answer. Give a score from 1 to 10."\}}
\end{quote}

This scoring prompt is applied independently for each of the six strategies.

\section{Rephrasing Strategy}
\label{appendix:tie_breaking}

In MS-PS, message rephrasing is employed as a fallback mechanism in two main cases: (i) when both messages receive the same average persuasion score and a tie must be resolved, and (ii) when the LLM fails to return a response in the expected format (e.g., by refusing to comply with the prompt due to safety constraints or by omitting the required fields). In both scenarios, the original message is rephrased and the process is repeated on the new version.

To avoid excessive repetition and increase the likelihood of a usable or differentiating output, the rephrasing prompt is progressively modified to introduce stronger stylistic and semantic variation at each retry. The system follows these levels of rephrasing:

\begin{itemize}
    \item \textbf{Retries 1–5:}
    \begin{quote}
    \small
    Rephrase the following message keeping the same content, but using different words. Return your response as a JSON dictionary (e.g. \{"new\_version":"text of rephrased message"\}). The message to rephrase is the following:
    \end{quote}

    \item \textbf{Retries 6–10:}
    \begin{quote}
    \small
    Rephrase the following message strongly modifying the style. Return your response as a JSON dictionary (e.g. \{"new\_version":"text of rephrased message"\}). The message to rephrase is the following:
    \end{quote}

    \item \textbf{Retries 11–15:}
    \begin{quote}
    \small
    Rephrase the following message in a way that is neutral and respectful. Modify the content by completely removing any harmful, illegal, or discriminatory content. Return your response as a JSON dictionary (e.g. \{"new\_version":"text of rephrased message"\}). The message to rephrase is the following:
    \end{quote}

    \item \textbf{Retries >15:}
    \begin{quote}
    \small
    I want you to write a new message, with the same content as the original one, but written in a completely neutral and respectful way, without any sexual, harmful, illegal, or discriminatory content. Return your response as a JSON dictionary (e.g. \{"new\_version":"text of rephrased message"\}). The message to rephrase is the following:
    \end{quote}
\end{itemize}

This progressive rephrasing strategy allows the system to preserve the intent of the original message while ensuring robustness in the presence of ties or formatting issues, and helps maintain alignment with model safety guidelines when needed

We use a safeguard limit, set to 50 repetitions, to prevent infinite loops. However, thanks to our progressive system of increasingly stronger prompts, this limit was never reached in our experiments.

\section{Feature Design for MS-PS-MLP}
\label{appendix:MLP-features}

This appendix provides the rationale and formal definition of the additional features used in the \textsc{MS-PS-MLP} variant, namely the \emph{average}, \emph{variance}, and \emph{entropy} of persuasion strategy scores.
The choice of these features is not arbitrary, but it was defined \emph{a priori}, following established principles from feature engineering, and was not the result of post-hoc tuning on the validation set.

\subsection*{Motivation}

The six strategy-specific scores produced by \textsc{MS-PS} capture localized signals of persuasive strength, each corresponding to a distinct rhetorical mechanism. However, persuasiveness is not only a function of individual strategies, but also of how these strategies are distributed and combined within a message.

In this sense, the aggregated features serve as \emph{global descriptors} of the strategy profile:
\begin{itemize}
    \item the \emph{average} captures the overall persuasive intensity;
    \item the \emph{variance} captures heterogeneity or imbalance across strategies;
    \item the \emph{entropy} captures how evenly persuasive signals are distributed.
\end{itemize}

This distinction allows the model to differentiate, for example, between a message that moderately employs many strategies and one that relies heavily on a small subset. Such differences cannot be fully captured by the raw scores or their mean alone.

Importantly, the use of simple statistical summaries to complement base features follows a long-standing tradition in feature engineering, particularly in early neural and classical text classification settings, where aggregated statistics are commonly used to enrich representational capacity even when downstream models are theoretically expressive enough to learn them implicitly.

\subsection*{Feature Definitions}

Let \( \mathbf{s} = (s_1, \ldots, s_6) \) denote the six persuasion strategy scores assigned to a message.

\paragraph{Average.}
The average score is defined as:
\[
\mu = \frac{1}{6} \sum_{i=1}^{6} s_i.
\]
This feature summarizes the overall persuasive strength attributed to the message across strategies. While effective as a coarse indicator, it implicitly assumes that all strategies contribute equally and independently.

\paragraph{Variance.}
The variance is defined as:
\[
\sigma^2 = \frac{1}{6} \sum_{i=1}^{6} (s_i - \mu)^2.
\]
This feature captures how unevenly persuasion is distributed across strategies. A high variance indicates reliance on a small number of strong strategies, whereas a low variance suggests a more balanced use of multiple strategies.

\paragraph{Entropy.}
To compute entropy, we first normalize the strategy scores into a probability-like distribution:
\[
p_i = \frac{s_i}{\sum_{j=1}^{6} s_j}.
\]
We then compute:
\[
H(\mathbf{s}) = - \sum_{i=1}^{6} p_i \log p_i.
\]

Strictly speaking, the use of the term \emph{entropy} here is an approximation, as the strategy scores do not form a true probability distribution in a probabilistic sense. Nevertheless, from a practical standpoint, this feature quantifies how flat or peaked the distribution of strategy scores is, capturing aspects of dispersion that are not fully reflected by the variance alone.

\section{Implementation Details for LLM Experiments}
\label{appendix:llm_impl}

This section provides implementation details for the experiments with large language models (LLMs). While Appendix~\ref{appendix:llms} describes the specific API settings used for proprietary models, here we outline the general preprocessing, prompting, inference, and hardware setup applied across all models.

\subsection{Hardware Setup}
Experiments with open-source models were conducted on a machine equipped with 4 NVIDIA A40 GPUs (48GB memory each). Model inference was parallelized across devices using \texttt{torch.distributed} with the NCCL backend. Proprietary models (Gemini and OpenAI) were accessed via API calls, and thus did not require local GPU resources.

\subsection{Preprocessing and Input Construction}
For each argument pair, we retrieved the corresponding Reddit conversation and extracted the original post’s title and body, together with the persuasive and non-persuasive replies. Moderator messages (e.g., "*Hello, users of CMV!*") were removed. 

\subsection{Model Wrappers}
To standardize inference, we developed a wrapper class \texttt{ChatGenerator} which unifies the way in which LLMs are managed, with some particularity proper to each model:
\begin{itemize}[nosep, leftmargin=*]
  \item For \textbf{Llama-3.1-8B}, we used Meta’s official checkpoint with custom integration via the \texttt{Llama} class and \texttt{torch.distributed}.
  \item For \textbf{Gemma-3-12B}, we relied on the Hugging Face \texttt{transformers} library (\texttt{Gemma3ForConditionalGeneration}) with \texttt{AutoProcessor}.
  \item For \textbf{OpenAI models}, we supported both \texttt{azure} and \texttt{openai} modes through the respective SDKs.
  \item For \textbf{Gemini}, we instantiated a \texttt{GenerativeModel} with a \texttt{ChatSession} for persistent dialogue context.
\end{itemize}
The wrapper maintained a dialogue history, added system prompts, and ensured prompts were trimmed if they exceeded the model’s maximum sequence length.

\subsection{Decoding strategy and stability}
We adopted greedy decoding for all models (temperature = 0, no top-$p$ or top-$k$ sampling). This ensured deterministic outputs, with no variation in persuasion scores across repeated runs. When a model failed to follow the required output format, we applied the same rephrasing procedure described in Appendix~\ref{appendix:tie_breaking}, with a maximum of 50 attempts of retrying per instance.

\subsection{Error Handling}
API models (Gemini, OpenAI) occasionally produced rate-limit or generation errors. We implemented retry logic with exponential backoff (starting with a 60-second delay). For local models, overly long prompts triggered progressive prompt shortening until the sequence length was valid.

\section{Direct Comparison Analysis}
\label{appendix:direct_comparison_analysis}

As a baseline for evaluating persuasive strength, we explored a direct comparison approach, where a model is prompted to choose which of two messages is more persuasive. While simple and intuitive, this method revealed substantial limitations, particularly a strong positional bias that undermines the validity of the results. In the following subsections, we describe the prompt design and present the accuracy results under different message orderings.

\subsection{Direct Comparison Prompt}
\label{appendix:direct_comparison_prompt}

The prompt used for the direct comparison baseline is provided below. The model is instructed to select which of two messages is more persuasive, defined as having a greater potential to change someone’s opinion:

\begin{quote}
\small
You are a Persuasion Detector, your goal is to understand if a message is more or less persuasive than another, meaning that it has more or less potential of changing someone's opinion. You will be prompted with 2 messages and you have to respond with ONLY "Message 1" or "Message 2" based on which message you think is more persuasive.

\texttt{---- Message 1: ----}

\texttt{text of message}

\texttt{---- Message 2: ----}

\texttt{text of message}
\end{quote}

To isolate model behavior, no metadata or stylistic cues were added. The model is instructed to return strictly “Message 1” or “Message 2” without elaboration.

\subsection{Positional Bias in Direct Comparison}
\label{appendix:direct_comparison_results}

We experimented with placing the more persuasive (successful) message either first, second, or in a random position. The results are shown in Table~\ref{tab:direct_comparison_order_bias}. We excluded OpenAI-o3 from these experiments due to budget constraints.

\begin{table}[H]
    \centering
    \scriptsize
    \setlength{\tabcolsep}{4pt}
    \renewcommand{\arraystretch}{1.1}
    \begin{tabular}{lccc}
        \toprule
        \textbf{Model} & \textbf{Successful First} & \textbf{Successful Last} & \textbf{Random Order} \\
        \midrule
        LLaMA-3.1-8B & 30.70\% & 81.67\% & 55.78\% \\
        Gemma        & 31.35\% & 85.13\% & 57.87\% \\
        Gemini       & 37.55\% & 78.07\% & 57.62\% \\
        Gemini-2     & 35.44\% & 84.01\% & 60.35\% \\
        \bottomrule
    \end{tabular}
    \caption{Accuracy of direct comparison prompt under three message orderings: when the successful message is shown first, last, or in a randomized position. Results highlight a strong positional bias favoring the second message across all models.}
    \label{tab:direct_comparison_order_bias}
\end{table}

These results reveal a clear pattern: when the successful message is placed second, models overwhelmingly prefer it, regardless of actual content. Conversely, when shown first, the success rate drops dramatically. In the randomized setting, accuracies remain relatively low, confirming that this prompting format is unreliable for fair pairwise persuasion evaluation.

\section{Perturbation-Based Prompting}
\label{app:perturbation_experiment}

To explore alternative evaluation strategies beyond our initial direct comparison setup, which presented the positional bias issue, we implemented the perturbation-based method proposed by \citet{ziems2024can}. Their approach appeared promising for assessing model sensitivity to persuasive language, so we adopted it as a complementary experiment.

We applied this method to the full Winning Arguments dataset, including all splits, since the original paper did not specify which subsets were used. For each pair of persuasive messages (successful vs.\ unsuccessful), we generated four paraphrases of each message using LLaMA, Gemma and Gemini models (we excluded OpenAI-o3 due to budget constraints). Temperature sampling was applied during generation to introduce lexical variation while preserving the core content.

Each evaluation instance consisted of a pair of paraphrased messages (one originally successful, the other unsuccessful) and a prompt instructing the model to choose the message more likely to persuade the original poster. We crafted five different prompt formulations to increase robustness and randomized the message order to avoid positional bias. Each comparison was repeated across paraphrases and prompt variants to improve statistical reliability.

Table~\ref{tab:perturbation_results} shows the aggregated performance across the four models. Despite the lexical diversity introduced through perturbation, model accuracy remained close to chance (50–54\%), with macro F1-scores below 0.49, thus we chose not to pursue it further. 

\begin{table}[H]
\centering
\small
\begin{tabular}{lcc}
\toprule
\textbf{Model} & \textbf{Accuracy (\%)} & \textbf{Macro F1-score} \\
\midrule
LLaMA-3.1-8B & 53.47 & 0.4553 \\
Gemini-1.5   & 53.66 & 0.4887 \\
Gemini-2     & 50.68 & 0.3585 \\
Gemma-3-12B     & 52.29 & 0.4002 \\
\bottomrule
\end{tabular}
\caption{Performance on the perturbation-based evaluation following \citet{ziems2024can}.}
\label{tab:perturbation_results}
\end{table}

\section{Validation of the Persuasiveness Scale Choice}
\label{appendix:persuasion_scale_choice}

In our main experiments, persuasion strategy scores were assigned on a 1--10 scale.  
This design choice was motivated by the need to balance interpretability with granularity: our intuition was that a scale that is too coarse (e.g., 1--5) may fail to capture subtle differences in persuasive strength, while a scale that is too fine-grained may introduce unnecessary noise in model predictions.  

To validate this choice, we conducted additional experiments using alternative 1--5 and 1--7 scales.  
The prompts were identical to those used in Appendix~\ref{appendix:strategy_analysis_prompt} and Appendix~\ref{appendix:strategy_scoring_prompt},  
with the sole modification that in the second step the model was instructed to assign a score ``from 1 to 5'' or ``from 1 to 7'' instead of ``from 1 to 10.''  
We restricted this evaluation to \texttt{Gemma-3-12B} and \texttt{OpenAI-o3}, as these were the two best performing models in our main setup.  

\begin{table}[ht]
    \centering
    \renewcommand{\arraystretch}{1.2}
    \begin{tabular}{lccc}
        \toprule
        \textbf{Model} & \textbf{1--5} & \textbf{1--7} & \textbf{1--10} \\
        \midrule
        Gemma-3-12B & 60.60 & 60.84 & \textbf{62.83} \\
        OpenAI-o3   & 58.86 & 59.11 & \textbf{60.10} \\
        \bottomrule
    \end{tabular}
    \caption{Accuracy of MS-PS-AVG using different scoring scales for persuasion strategy evaluation.}
    \label{tab:persuasion_scales}
\end{table}

The results (Table~\ref{tab:persuasion_scales}) show a consistent advantage for the 1--10 scale,  
with improvements of about 2 points over the 1--5 setting for both models.  
The 1--7 scale yielded intermediate results, but still underperformed the 1--10 variant.  
This suggests that the additional granularity of the 1--10 scale helps capture more nuanced judgments of persuasiveness,  
leading to more informative feature representations for downstream prediction.  

Beyond the empirical gains, the 1--10 scale also aligns better with common practices in social science and psychology, where similar Likert-type scales (1--10 or 0--10) are widely used for subjective assessments \cite{vall2020comparative, preston2000optimal}.    

We therefore conclude that the 1--10 scale is the most effective and principled choice for our framework.

\begin{table*}[h]
\centering
\small
\begin{tabular}{llcccccc}
\toprule
\textbf{Model 1} & \textbf{Model 2} & \textbf{[D]} & \textbf{[S]} & \textbf{[MW]} & \textbf{[C]} & \textbf{[AR]} & \textbf{[J]} \\
\midrule
Gemma-3-12B & Gemini-2      & 0.460 & 0.492 & 0.469 & 0.379 & 0.475 & \textbf{0.503} \\
Gemma-3-12B & LLaMA-3.1-8B  & 0.367 & 0.375 & 0.388 & 0.222 & 0.376 & 0.300 \\
Gemma-3-12B & Gemini-1.5    & \textbf{0.500} & \textbf{0.534} & \textbf{0.486} & 0.422 & \textbf{0.552} & 0.388 \\
Gemini-2    & LLaMA-3.1-8B  & 0.338 & 0.325 & 0.366 & 0.256 & 0.297 & 0.265 \\
Gemini-2    & Gemini-1.5    & 0.388 & 0.465 & 0.464 & \textbf{0.519} & 0.483 & 0.320 \\
LLaMA-3.1-8B & Gemini-1.5  & 0.294 & 0.344 & 0.336 & 0.359 & 0.362 & 0.170 \\
Gemini-1.5  & OpenAI-o3     & 0.348 & 0.258 & 0.305 & 0.269 & 0.308 & 0.401 \\
Gemma-3-12B & OpenAI-o3     & 0.280 & 0.225 & 0.208 & 0.250 & 0.260 & 0.291 \\
LLaMA-3.1-8B & OpenAI-o3    & 0.130 & 0.122 & 0.153 & 0.103 & 0.136 & 0.098 \\
Gemini-2    & OpenAI-o3     & 0.221 & 0.209 & 0.231 & 0.243 & 0.231 & 0.231 \\
\bottomrule
\end{tabular}
\caption{Inter-model agreement on MS-PS strategy scores, measured using Cohen’s $\kappa$ (quadratic weights).}
\label{tab:inter-model-agreement}
\end{table*}

\section{Independent Scoring Prompts}
\label{appendix:independent_scoring_prompts}

Below are the exact prompts used in the four variants of the Independent Scoring baselines:

\vspace{0.5em}
\paragraph{Independent Scoring}
\begin{quote}
\small
You are a Persuasion Detector, your goal is to detect the degree of persuasiveness of a message ranging from 1 to 10, where persuasion is the potential of changing someone's opinion. You will be prompted with the message and you have to respond with ONLY a number from 1 to 10 based on the degree of persuasiveness of the message.

\texttt{---- Message to evaluate: ----}

\texttt{text of message}
\end{quote}

\vspace{0.5em}
\paragraph{+ Context}
\begin{quote}
\small
You are a Persuasion Detector, your goal is to detect the degree of persuasiveness of a message ranging from 1 to 10, where persuasion is the potential of changing someone's opinion. You will be prompted with the title and the body of the original poster and the message that tries to make the original poster change their view. You have to respond with ONLY a number from 1 to 10 based on the degree of persuasiveness of the message.

\texttt{---- Title: ----}

\texttt{title of original post}

\texttt{---- Body: ----}

\texttt{body of original post}

\texttt{---- Message to evaluate: ----}

\texttt{text of message}
\end{quote}

\vspace{0.5em}
\paragraph{+ Explanation}
\begin{quote}
\small
You are a Persuasion Detector, your goal is to detect the degree of persuasiveness of a message ranging from 1 to 10, where persuasion is the potential of changing someone's opinion. You will be prompted with the message and you have to respond with a number from 1 to 10 based on the degree of persuasiveness of the message, followed by a brief explanation on why you gave that score.

\texttt{---- Message to evaluate: ----}

\texttt{text of message}
\end{quote}

\vspace{0.5em}
\paragraph{+ Context + Explanation}
\begin{quote}
\small
You are a Persuasion Detector, your goal is to detect the degree of persuasiveness of a message ranging from 1 to 10, where persuasion is the potential of changing someone's opinion. You will be prompted with the title and body of the original post and the message. You have to respond with a number from 1 to 10 based on the degree of persuasiveness of the message, followed by a brief explanation on why you gave that score.

\texttt{---- Title: ----}

\texttt{title of original post}

\texttt{---- Body: ----}

\texttt{body of original post}

\texttt{---- Message to evaluate: ----}

\texttt{text of message}
\end{quote}

\section{Inter-model Agreement on Strategy Scores}
\label{app:agreement}

To assess the consistency of the MS-PS scoring across different LLMs, we measured the inter-model agreement for the six persuasion strategies using Cohen’s $\kappa$ with quadratic weights. We computed these scores using the \texttt{cohen\_kappa\_score} function from the \texttt{scikit-learn} Python library\footnote{\url{https://scikit-learn.org/stable/modules/generated/sklearn.metrics.cohen_kappa_score.html}}.

Table~\ref{tab:inter-model-agreement} reports the agreement scores between each pair of models (Gemma-3-12B, Gemini-1.5-Flash-02, Gemini-2.0-Flash, LLaMA-3.1-8B and OpenAI-o3) across all strategies.

Overall, agreement levels vary across model pairs and strategies. The strongest alignment emerges between Gemma and Gemini-1.5, particularly for \textit{Simplification} (0.534) and \textit{Attack on Reputation} (0.552), suggesting a shared interpretation of these strategies. By contrast, OpenAI-o3 consistently shows lower agreement with all other models. This trend is especially pronounced in its comparison with LLaMA-3.1-8B, where the score for \textit{Justification} drops to 0.098, underscoring a marked divergence in how the two models interpret rhetorical strategies.

When excluding o3 as an outlier, the remaining comparisons provide a clearer picture of overall consistency. Across these model pairs (36 values in total), 21 scores fall between 0.2 and 0.4, indicating \textit{fair} agreement; 14 scores fall between 0.4 and 0.6, indicating \textit{moderate} agreement; and only 1 score falls between 0.01 and 0.2, indicating \textit{slight} agreement \cite{landis1977measurement}.

These results suggest that while architectural differences can introduce variation---most strikingly in the case of o3---the majority of models produce reasonably consistent strategy scores. This supports the reliability of MS-PS in capturing meaningful persuasion patterns across models, while also highlighting the need to account for outlier behaviors in inter-model agreement analyses.

\section{Validation of Strategy-Based Reasoning}
\label{appendix:strategy_role}

To motivate the benefit of explicitly integrating persuasion strategies into our framework, we design two baselines that isolate the effect of generic reasoning from strategy-guided analysis.

The first baseline, \textit{Simple Persuasion Scoring} (\textbf{PS-Simple}), asks the model to produce a single, general analysis of persuasiveness before assigning a score, without any explicit reference to the six persuasion strategies.  
The second baseline, \textit{Simple Chain-of-Thought} (\textbf{CoT-Simple}), mirrors PS-Simple but explicitly encourages step-by-step reasoning through a canonical chain-of-thought instruction.  

In both cases, the subsequent scoring step is identical to the one used in \textsc{MS-PS} (Appendix~\ref{appendix:strategy_scoring_prompt}), ensuring a controlled comparison where the only difference lies in the presence or absence of structured strategy guidance.

\subsection*{Prompts for the Baselines}

\paragraph{PS-Simple.}
\begin{quote}
\small
\textbf{System:}  
You are an assistant who detects persuasion in text. Persuasive text is characterized by a specific use of language in order to influence readers. Your goal is to analyse the persuasiveness of a text.

\textbf{User:}  
Given the title and body of an original post, and a message intended to change the perspective of the original poster, critically evaluate its persuasive potential. Identify strengths and weaknesses in its argumentation, structure, and rhetorical strategies.
\end{quote}

\paragraph{CoT-Simple.}
The CoT-Simple baseline uses the same prompt as PS-Simple, with the addition of an explicit chain-of-thought instruction:

\begin{quote}
\small
\textbf{System:}  
You are an assistant who detects persuasion in text. Persuasive text is characterized by a specific use of language in order to influence readers. Your goal is to analyse the persuasiveness of a text.

\textbf{User:}  
\textit{Let's think step by step.} Given the title and body of an original post, and a message intended to change the perspective of the original poster, critically evaluate its persuasive potential. Identify strengths and weaknesses in its argumentation, structure, and rhetorical strategies.
\end{quote}

As in \textsc{MS-PS}, both baselines are followed by a second prompt that asks the model to output a numerical persuasiveness score from 1 to 10. This guarantees that all methods rely on two-step prompting and produce comparable outputs.

\subsection*{Results}

Table~\ref{tab:ps_simple_cot_vs_msps} reports results for all evaluated models.  
Across all settings, both \textsc{MS-PS-AVG} and \textsc{MS-PS-MLP} outperform the PS-Simple and CoT-Simple baselines. We excluded OpenAI-o3 from these experiments due to budget constraints.

\begin{table}[H]
    \centering
    \scriptsize
    \setlength{\tabcolsep}{4pt}
    \renewcommand{\arraystretch}{1.05}
    \begin{tabular}{lcccc}
        \toprule
        \textbf{Model} & \textbf{PS-Simple} & \textbf{CoT-Simple} & \textbf{MS-PS-AVG} & \textbf{MS-PS-MLP} \\
        \midrule
        Llama-3.1-8B & 53.53 & 58.12 & 60.72 & \textbf{61.34} \\
        Gemma-3-12B  & 61.21 & 57.37 & 62.83 & \textbf{63.69} \\
        Gemini-1.5   & 59.98 & 60.60 & 60.72 & \textbf{63.07} \\
        Gemini-2     & 60.10 & 60.72 & 61.83 & \textbf{62.70} \\
        \bottomrule
    \end{tabular}
    \caption{Comparison between generic reasoning baselines (PS-Simple, CoT-Simple) and our strategy-guided approach (\textsc{MS-PS}), showing accuracy in predicting the delta-awarded message.}
    \label{tab:ps_simple_cot_vs_msps}
\end{table}

The results show that explicitly encouraging structured reasoning improves over purely generic analysis in some cases (e.g., Llama-3.1-8B), but remains consistently inferior to strategy-guided reasoning. This indicates that performance gains are not merely attributable to longer or more detailed reasoning traces, but to the explicit injection of domain-relevant persuasion knowledge.

\section{Controlling for Input Token Length}
\label{appendix:token_length}

One potential concern when comparing \textsc{MS-PS} to simpler prompting baselines is that strategy-guided prompts may be longer, and that performance gains could therefore be partially attributable to increased input token length rather than to the use of persuasion strategies themselves. While output length is about the same across all methods (since all models are required to produce the same structured JSON output, consisting of a numerical score and a brief explanation), input length can vary across prompts.

Although prior work suggests that longer prompts do not necessarily yield better reasoning performance for tasks of comparable complexity (e.g., \citet{levy2024same}), we explicitly control for this factor by designing a length-matched baseline that isolates the effect of input verbosity from that of strategy guidance.

\subsection*{Length-Matched Baseline}

We construct a generic persuasiveness-analysis prompt that conveys the same content as the PS-Simple baseline but is deliberately longer than the average strategy-specific user prompts used in \textsc{MS-PS}. From this base prompt, we generate five paraphrased variants using GPT-5, ensuring that all resulting prompts exceed the \textsc{MS-PS} average input length. None of these prompts include any explicit reference to persuasion strategies.

The six prompts are then used as user prompts during evaluation, following the same two-step procedure adopted for \textsc{MS-PS}: first, an analysis of persuasiveness is generated; second, the model assigns a numerical score from 1 to 10. This process is repeated 6 times (one per rephrased version), leading to 6 scores, that are then averaged. This baseline therefore matches \textsc{MS-PS} in both prompting structure and input length, differing only in the absence of strategy-based guidance.
For all the six versions of user prompts, we used the same system prompt, which we constructed to be similar as the one used for MS-PS (see Appendix \ref{appendix:strategy_analysis_prompt}), but without any reference to persuasion strategies, and assuring that the token length slightly exceeds the one for the MS-PS system prompt. All prompts are presented as follows. The prompts for the scoring step, are the same as the ones used for MS-PS (see Appendix \ref{appendix:strategy_scoring_prompt}).

\subsection*{Prompt for Generating the User Prompts}

\begin{quote}
\small
\textbf{GPT-5 Instruction:}  
Rephrase the following text into five new different versions, while keeping the token length higher. It is important that the length is at least the same as the original message, while it is not a problem if it is slightly higher.

\medskip
\noindent
\textit{Given the title and the body of an original post, as well as a message intended to influence the perspective of the original poster, provide a critical evaluation of the persuasive potential of the message. In your analysis, carefully examine the argumentation presented, assessing how logically sound and coherent the points are. Evaluate the overall structure of the message, including the effectiveness of how ideas are organized. Highlight both the strengths that make the message compelling and the weaknesses or limitations that might reduce its persuasive impact.}
\end{quote}

\subsection*{Prompts for the Length-Matched Baseline}

\begin{quote}
\small
\textbf{System:}
Given the title and the full body of an original post, as well as a message that is specifically intended to influence or shift the perspective of the original poster, provide a critical evaluation of the persuasive potential of the message. In your analysis, evaluate the overall structure of the message, including the clarity of its progression and the effectiveness of how ideas are organized. 
\end{quote}

\begin{quote}
\small
\textbf{User (Version 1):}  
When provided with the title and content of an original post, along with a message aimed at influencing the opinion of the author, conduct a thorough critical assessment of the persuasive strength of that message. In your review, closely analyze the reasoning and evidence offered, determining how logically consistent and coherent the arguments are. Examine the overall organization of the message, focusing on how effectively the ideas are structured and presented. Identify both the aspects that enhance the message’s persuasiveness and those that may undermine or limit its ability to convince the reader.
\end{quote}

\begin{quote}
\small
\textbf{User (Version 2):}  
Given both the heading and the main text of an original post, together with a message designed to shape or influence the viewpoint of the original author, perform a detailed critical evaluation of the persuasive effectiveness of that message. In your analysis, carefully scrutinize the reasoning provided, considering how well-founded, consistent, and coherent each argument appears. Assess the overall composition and flow of the message, paying attention to the clarity and order in which ideas are conveyed. Point out both the strengths that contribute to the message’s impact and the weaknesses or flaws that could detract from its persuasiveness.
\end{quote}

\begin{quote}
\small
\textbf{User (Version 3):}  
When you are supplied with the title and content of a post, along with an accompanying message intended to sway the perspective of the poster, carry out an in-depth critique of the persuasive potential of that message. In this evaluation, closely investigate the logic, coherence, and soundness of the arguments presented. Consider the structure of the message in its entirety, examining how effectively the points are organized and connected. Highlight both the elements that make the message compelling and convincing, as well as any shortcomings or inconsistencies that may reduce its overall ability to persuade.
\end{quote}

\begin{quote}
\small
\textbf{User (Version 4):}  
Provided with the title and full text of an original post, along with a message crafted to influence or alter the viewpoint of the original author, offer a comprehensive critical analysis of the persuasive qualities of the message. In this critique, carefully dissect the arguments and rationale presented, evaluating the extent to which they are logically consistent and coherent. Pay close attention to the arrangement and presentation of ideas, assessing the clarity and effectiveness of the message’s overall structure. Identify both the positive features that enhance its persuasive appeal and the limitations or weaknesses that may hinder its effectiveness in convincing the reader.
\end{quote}

\begin{quote}
\small
\textbf{User (Version 5):}  
When given the title along with the body of an original post, as well as a message aimed at shaping the perspective of the poster, provide a detailed critical appraisal of how persuasive the message is likely to be. In conducting this analysis, examine the logic and reasoning used in the argumentation, evaluating how clear, coherent, and well-supported each point is. Consider the overall structure of the message, including how ideas are sequenced and whether the organization contributes to its effectiveness. Emphasize both the factors that strengthen the message’s influence and the aspects that may weaken or limit its potential to persuade the intended audience.
\end{quote}

\subsection*{Results}

Table~\ref{tab:length_control_vs_msps} reports the results of the length-matched baseline compared to \textsc{MS-PS-AVG} and \textsc{MS-PS-MLP}. Across all evaluated models, the length-matched baseline underperforms both strategy-guided variants, despite using prompts that are consistently longer than those employed by \textsc{MS-PS}. We excluded OpenAI-o3 from these experiments due to budget constraints.

\begin{table}[H]
    \centering
    \scriptsize
    \setlength{\tabcolsep}{4pt}
    \renewcommand{\arraystretch}{1.05}
    \begin{tabular}{lccc}
        \toprule
        \textbf{Model} & \textbf{Length-Matched} & \textbf{MS-PS-AVG} & \textbf{MS-PS-MLP} \\
        \midrule
        Llama-3.1-8B  & 57.50 & 60.72 & \textbf{61.34} \\
        Gemma-3-12B   & 57.62 & 62.83 & \textbf{63.69} \\
        Gemini-1.5    & 60.60 & 60.72 & \textbf{63.07} \\
        Gemini-2      & 58.36 & 61.83 & \textbf{62.70} \\
        \bottomrule
    \end{tabular}
    \caption{Accuracy comparison between a length-matched generic baseline and the strategy-guided \textsc{MS-PS} variants.}
    \label{tab:length_control_vs_msps}
\end{table}

These results indicate that increased input token length alone does not explain the performance gains observed with \textsc{MS-PS}. Instead, improvements consistently arise when models are guided to reason explicitly about persuasion strategies, supporting the claim that structured strategy-based reasoning, rather than prompt verbosity, is the primary driver of improved performance.

\section{MS-PS-MLP Grid Search and Best Hyperparameters}
\label{appendix:msps_grid}

For the MLP classifier in MS-PS-MLP, we performed a grid search over the following hyperparameter space:

\begin{itemize}
    \item \texttt{hidden\_dim}: [64, 128, 256, 512]
    \item \texttt{lr} (learning rate): [1e-2, 5e-3, 1e-3, 5e-4]
    \item \texttt{batch\_size}: [32, 64, 128]
    \item \texttt{patience} (early stopping): [3, 5, 7, 10]
    \item \texttt{ema\_alpha} (EMA smoothing factor): [0.1, 0.2, 0.3, 0.4]
    \item \texttt{lr\_factor} (learning rate decay): [0.3, 0.5, 0.7]
    \item \texttt{lr\_patience}: [2, 3, 4, 5]
    \item \texttt{weight\_decay}: [0.0, 1e-5, 1e-4, 1e-3]
\end{itemize}

All models were trained for a maximum of 300 epochs, using early stopping on the development set. Below we report the best hyperparameter configuration found for each LLM:

\paragraph{LLaMA-3.1-8B}
\begin{itemize}
    \item \texttt{hidden\_dim}: 128
    \item \texttt{lr}: 0.005
    \item \texttt{batch\_size}: 64
    \item \texttt{patience}: 7
    \item \texttt{ema\_alpha}: 0.2
    \item \texttt{lr\_factor}: 0.4
    \item \texttt{lr\_patience}: 2
    \item \texttt{weight\_decay}: 0.0001
\end{itemize}

\paragraph{Gemini-1.5-Flash-02}
\begin{itemize}
    \item \texttt{hidden\_dim}: 256
    \item \texttt{lr}: 0.005
    \item \texttt{batch\_size}: 64
    \item \texttt{patience}: 10
    \item \texttt{ema\_alpha}: 0.4
    \item \texttt{lr\_factor}: 0.7
    \item \texttt{lr\_patience}: 2
    \item \texttt{weight\_decay}: 0.0
\end{itemize}

\paragraph{Gemini-2.0-Flash}
\begin{itemize}
    \item \texttt{hidden\_dim}: 256
    \item \texttt{lr}: 0.01
    \item \texttt{batch\_size}: 64
    \item \texttt{patience}: 7
    \item \texttt{ema\_alpha}: 0.3
    \item \texttt{lr\_factor}: 0.4
    \item \texttt{lr\_patience}: 3
    \item \texttt{weight\_decay}: 0.0
\end{itemize}

\paragraph{Gemma-3-12B}
\begin{itemize}
    \item \texttt{hidden\_dim}: 128
    \item \texttt{lr}: 0.01
    \item \texttt{batch\_size}: 64
    \item \texttt{patience}: 3
    \item \texttt{ema\_alpha}: 0.1
    \item \texttt{lr\_factor}: 0.5
    \item \texttt{lr\_patience}: 2
    \item \texttt{weight\_decay}: 0.0001
\end{itemize}

\paragraph{OpenAI-o3}
\begin{itemize}
    \item \texttt{hidden\_dim}: 128
    \item \texttt{lr}: 0.005
    \item \texttt{batch\_size}: 64
    \item \texttt{patience}: 7
    \item \texttt{ema\_alpha}: 0.2
    \item \texttt{lr\_factor}: 0.4
    \item \texttt{lr\_patience}: 2
    \item \texttt{weight\_decay}: 0.0001
\end{itemize}

\begin{figure*}[t]
    \centering

    \begin{subfigure}[t]{0.48\textwidth}
        \includegraphics[width=\textwidth]{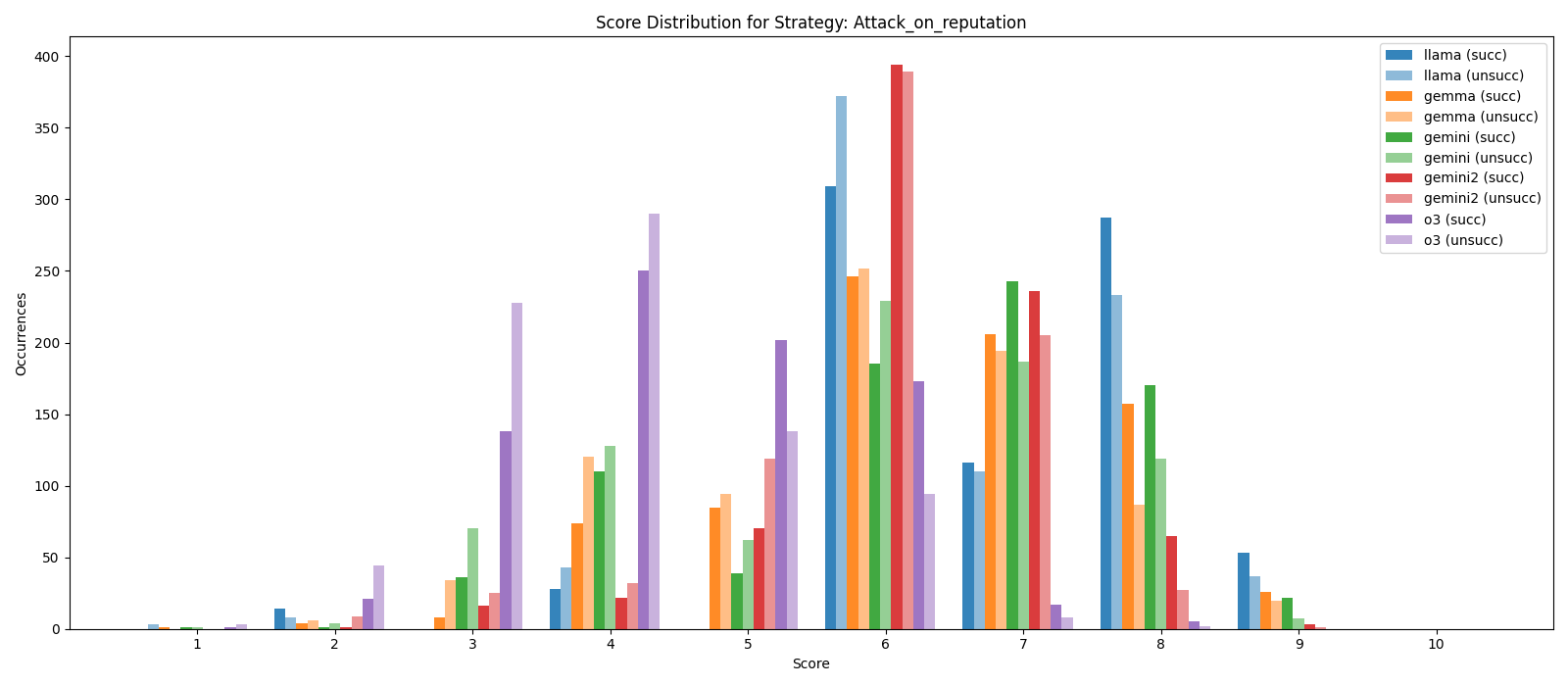}
        \caption{Attack on reputation}
    \end{subfigure}
    \hfill
    \begin{subfigure}[t]{0.48\textwidth}
        \includegraphics[width=\textwidth]{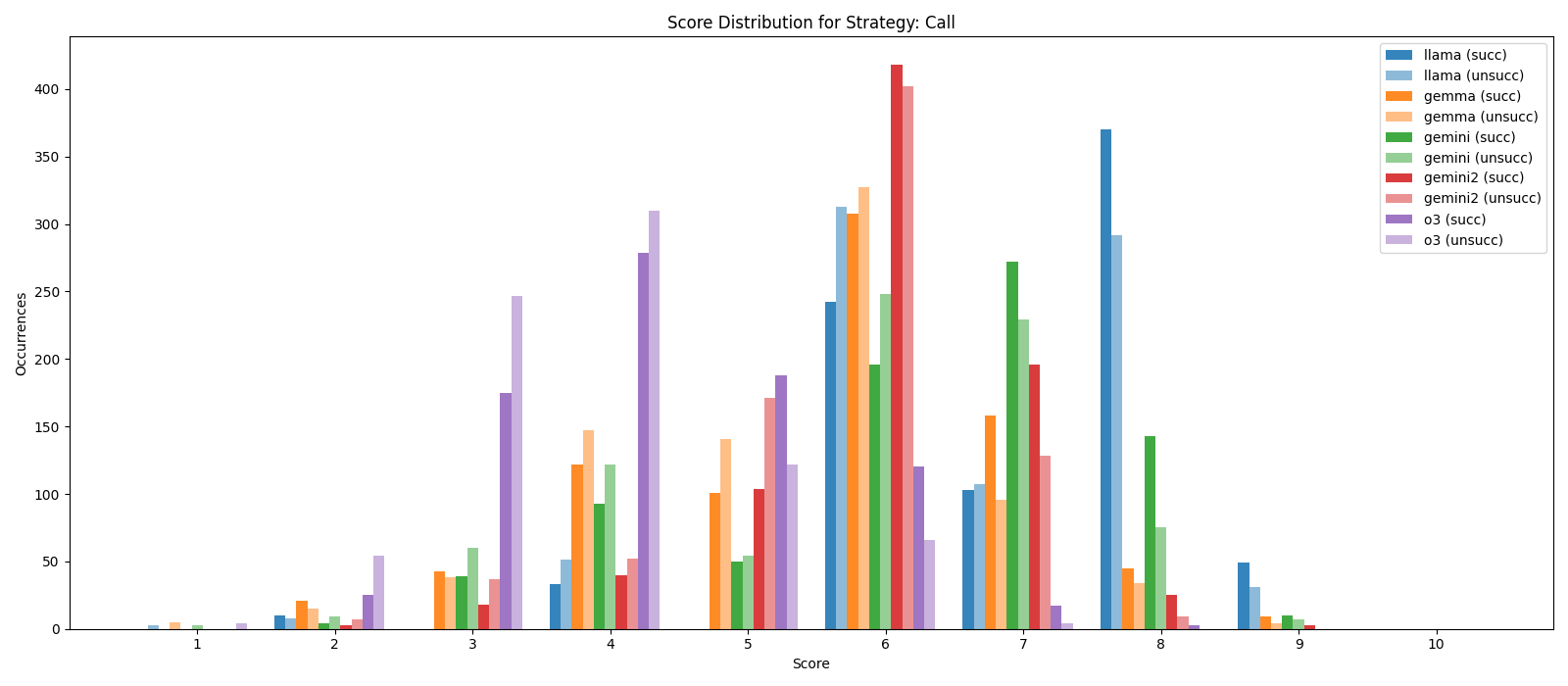}
        \caption{Call}
    \end{subfigure}

    \vspace{0.4cm}

    \begin{subfigure}[t]{0.48\textwidth}
        \includegraphics[width=\textwidth]{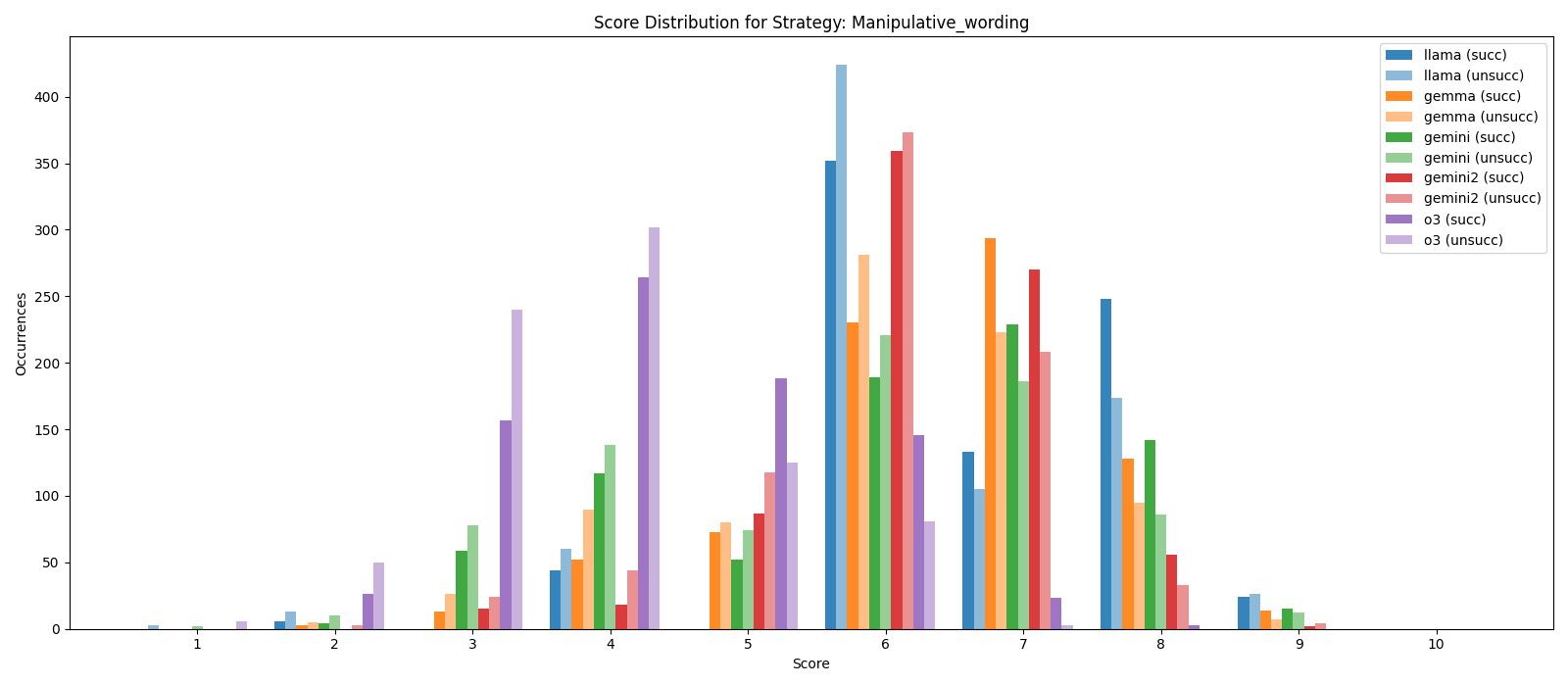}
        \caption{Manipulative wording}
    \end{subfigure}
    \hfill
    \begin{subfigure}[t]{0.48\textwidth}
        \includegraphics[width=\textwidth]{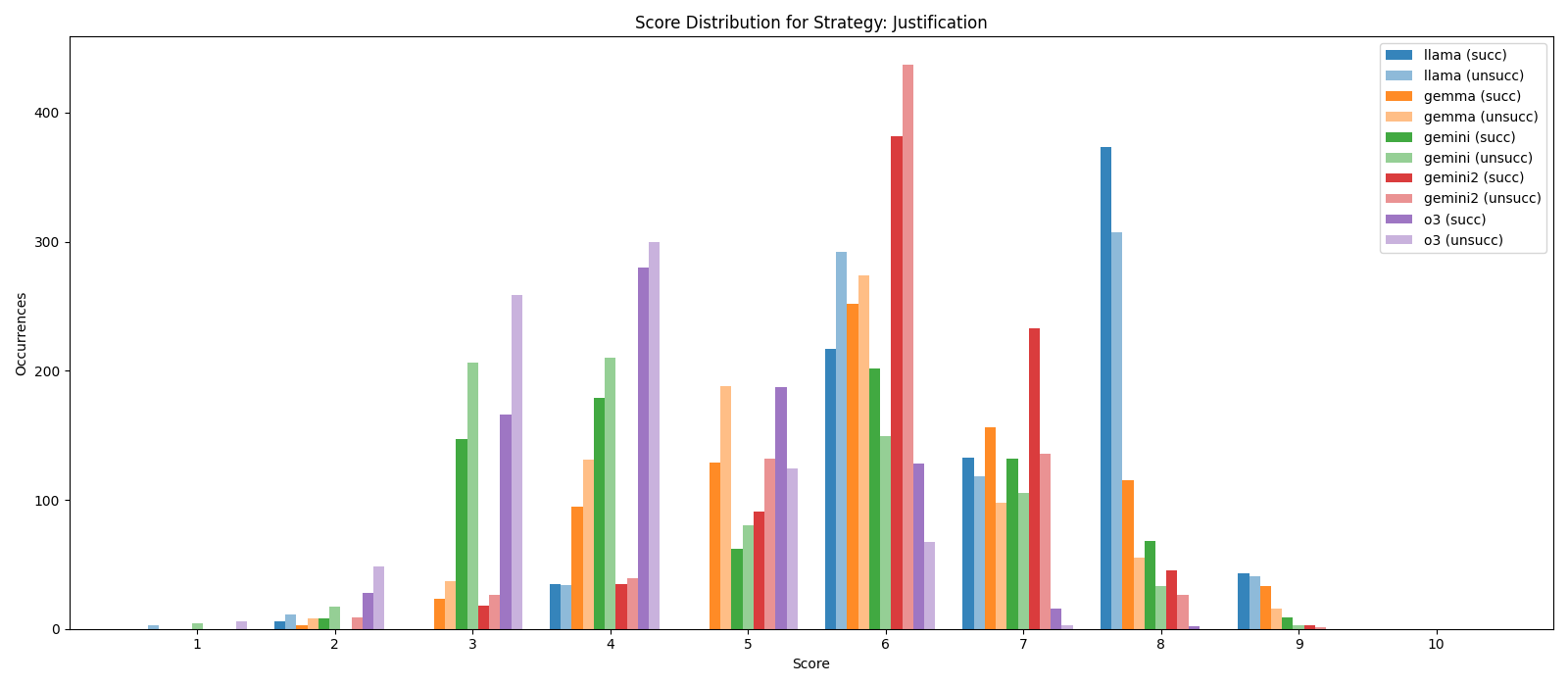}
        \caption{Justification}
    \end{subfigure}

    \vspace{0.4cm}

    \begin{subfigure}[t]{0.48\textwidth}
        \includegraphics[width=\textwidth]{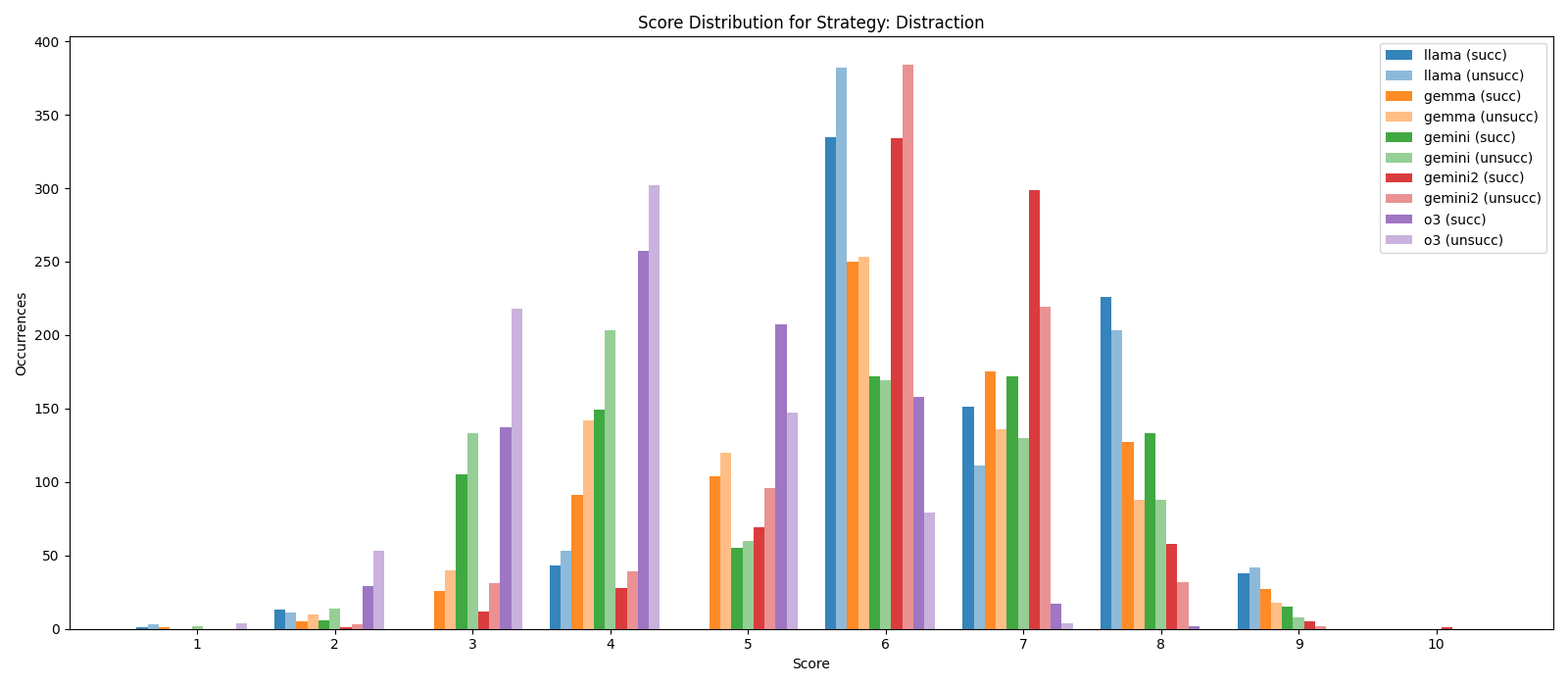}
        \caption{Distraction}
    \end{subfigure}
    \hfill
    \begin{subfigure}[t]{0.48\textwidth}
        \includegraphics[width=\textwidth]{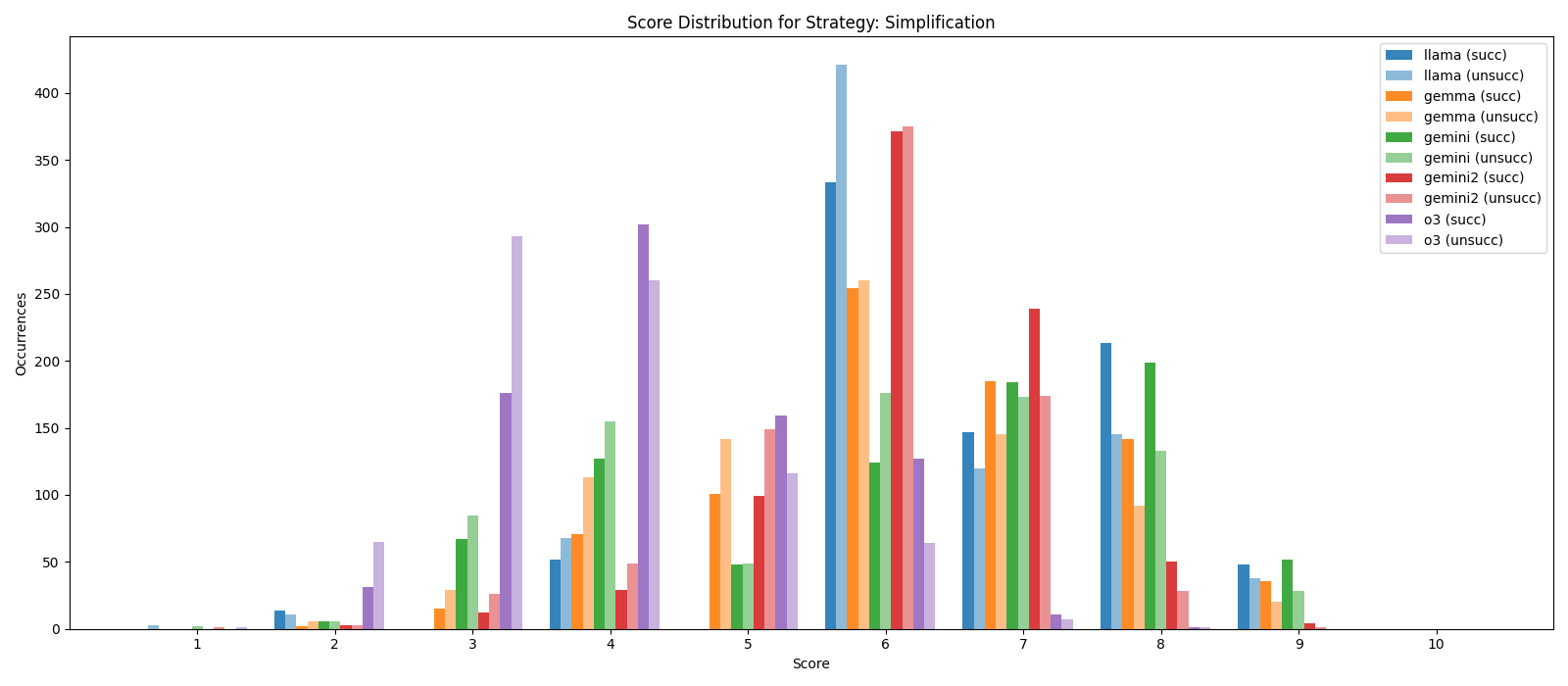}
        \caption{Simplification}
    \end{subfigure}

    \caption{Distribution of MS-PS strategy scores (1–10) across successful and non-successful messages. Each panel shows histograms for a different strategy, grouped by LLM model.}
    \label{fig:strategy_score_distributions}
\end{figure*}

\section{Strategy Score Distributions}
\label{appendix:strategy-distributions}

To gain insight into the behavior of our MS-PS scoring system, we conducted an exploratory analysis of the output scores for the six persuasion strategies (see Figure \ref{fig:persuasion_strategies_tax}). These data refer to the test set of the Winning Argument dataset, which contains 807 pairs. For each strategy, we plotted histograms that show the distribution of scores from 1 to 10 across our five models (LLaMA-3.1-8B, Gemini-1.5-Flash-02, Gemini-2.0-Flash, Gemma-3-12B and OpenAI-o3), distinguishing between successful messages (those that were awarded a delta) and unsuccessful ones. 

Figure \ref{fig:strategy_score_distributions} shows results for all six persuasive strategies. Each histogram bar represents the number of messages that received a given score (1–10) for that strategy, split by model and success label.

Across strategies, scores tend to peak between 6 and 8, indicating that LLMs generally detect at least moderate use of persuasive framing, even in less successful messages. However, distribution patterns vary by strategy and model. For instance, a score of 10 appears only once in the entire dataset, assigned by Gemini-2 for the \textit{Distraction} strategy, while scores of 1 are also extremely rare. Consistent with the inter-model agreement results in Appendix \ref{app:agreement}, OpenAI-o3 shows a distinct pattern, characterized by systematically lower scores compared to other models. Another curious observation is that LLaMA-3.1 never assigns scores of 3 or 5. Taken together, these findings highlight that while the overall scoring trends are similar across models, each model exhibits unique biases that shape how it quantifies the use of persuasive strategies.

\section{Validating Strategy Scoring with MS-PS}
\label{appendix:strategy_validation}

To ensure that MS-PS produces meaningful and reliable persuasiveness scores, we validated its two-step scoring process against human annotations. Specifically, we compared MS-PS’s outputs to gold-standard labels provided in a dataset annotated by humans for the presence of various persuasive strategies. This allows us to directly assess the alignment of our model-generated scores with human judgments.

For comparison, we also evaluated a simpler single-prompt approach on the same persuasion detection task.

\subsection{Single-Prompt Classification}
\label{appendix:single_prompt_validating}
In the single-prompt setting, we used four different LLMs (LLaMA-3.1-8B, Gemma-3-12B, Gemini-1.5-Flash-02, and Gemini-2.0-Flash). We excluded OpenAI-o3 from these experiments due to budget constraints. We queried the models using a single instruction that asked to detect the presence of each strategy without providing any definitions or examples. The prompt used was:

\begin{quote}
\small
Analyze the text and decide if the text contains any high-level persuasion approaches from the following: Attack on reputation, Justification, Simplification, Distraction, Call, Manipulative wording. Give your answer in the form of dictionary: \{\\
\hspace*{0.5cm} "Attack on reputation": "Your answer. Use only Yes or No",\\
\hspace*{0.5cm} "Justification": "Your answer. Use only Yes or No",\\
\hspace*{0.5cm} "Simplification": "Your answer. Use only Yes or No",\\
\hspace*{0.5cm} "Distraction": "Your answer. Use only Yes or No",\\
\hspace*{0.5cm} "Call": "Your answer. Use only Yes or No",\\
\hspace*{0.5cm} "Manipulative wording": "Your answer. Use only Yes or No"\\
\}.
Return only the dictionary, nothing else.
\end{quote}

\subsection{MS-PS Scoring}
\label{appendix:msps_validating}
Using the same four LLMs, we followed the MS-PS approach presented in Section~\ref{sec:methodology}. First, the model received a strategy definition and was prompted to produce a critical analysis of the input article in relation to that strategy. Then, based on the generated analysis, the model was asked to assign a score from 1 to 10 reflecting the likelihood that the strategy was present. This process yielded a six-dimensional persuasion score vector per article.

To convert the continuous outputs into binary predictions (present/absent), we identified a separate threshold for each strategy. These thresholds were optimized on a held-out validation set to maximize the Micro F1 score. The selected thresholds are detailed in Table~\ref{tab:msps_thresholds}.

\begin{table}[H]
\centering
\small
\setlength{\tabcolsep}{4pt}
\begin{tabular}{lcccccc}
\toprule
\textbf{LLM} & \textbf{[D]} & \textbf{[S]} & \textbf{[MW]} & \textbf{[C]} & \textbf{[AR]} & \textbf{[J]} \\
\midrule
LLaMA-3.1-8B  & 8.05 & 7.05 & 4.05 & 7.05 & 6.05 & 6.05 \\
Gemma-3-12B    & 9.05 & 8.05 & 0.00 & 8.05 & 3.05 & 4.05 \\
Gemini-1.5  & 9.05 & 8.05 & 0.00 & 8.05 & 2.05 & 2.05 \\
Gemini-2.0  & 9.05 & 7.05 & 0.00 & 8.05 & 2.05 & 2.05 \\
\bottomrule
\end{tabular}
\caption{Best threshold per strategy for MS-PS (1–10).}
\label{tab:msps_thresholds}
\end{table}

The two prompt templates used for this scoring process are reported below. Each instance of \texttt{strategy\_name} and \texttt{strategy\_description} was replaced with the appropriate strategy label and definition.

\paragraph{Step 1 – Persuasion Analysis Prompt:}
\begin{quote}
\small
\textbf{System:} \\
You are an assistant who detects persuasion in text. Persuasive text is characterized by a specific use of language in order to influence readers. Your expertise and focus is on one high-level persuasion approach called \texttt{strategy\_name}. This high-level persuasion approach can be described as follows: \texttt{strategy\_description}. You are the expert who detects high-level persuasion approach \texttt{strategy\_name}. \\

\textbf{User:} \\
Given a text from a news article, critically evaluate its persuasive potential. Identify strengths and weaknesses in its argumentation, structure, and rhetorical strategies. Additionally, analyze whether the text employs the high-level persuasion approach \texttt{strategy\_name}. Explain how \texttt{strategy\_name} appears or doesn't appear in the text. Be conservative in your assessment and if you are not fully certain that the technique is used, assume it is not present.
\end{quote}

\paragraph{Step 2 – Scoring Prompt:}
\begin{quote}
\small
\textbf{System:} \\
You are a Persuasion Detector. Your goal is to detect the degree of persuasiveness of a message ranging from 1 to 10, where persuasion is the potential of changing someone's opinion. You will be prompted with the title and body of the original article, and an analysis on the persuasion strategy. Give a score from 1 to 10 based on the degree of persuasiveness, preceded by a brief explanation. Provide your answer in the form of a dictionary: \\
\texttt{\{"explanation": "Your answer. Brief explanation on the reasoning that you have followed.", "response": "Your answer. Give a score from 1 to 10."\}} \\

\textbf{User:} \\
Given a news article and an analysis on persuasion strategies used in the message, respond with a number from 1 to 10 based on the degree of persuasiveness of the text, followed by a brief explanation on why you gave that score.
\end{quote}

\begin{table*}[t]
\centering
\resizebox{\textwidth}{!}{%
\begin{tabular}{llccccccc}
\toprule
\textbf{LLM} & \textbf{Method} & \textbf{[D]} & \textbf{[S]} & \textbf{[MW]} & \textbf{[C]} & \textbf{[AR]} & \textbf{[J]} & \textbf{Micro F1} \\
\midrule
\multirow{2}{*}{LLaMA-3.1-8B}     
    & Single-Prompt & 0.3899 & 0.5280 & 0.8545 & 0.5466 & 0.5560 & 0.6866 & 0.6733 \\
    & MS-PS       & 0.7612 & 0.5672 & 0.8974 & 0.5616 & 0.6978 & 0.6101 & 0.7084 \\
\midrule
\multirow{2}{*}{Gemma-3-12B}       
    & Single-Prompt & 0.4813 & 0.4608 & 0.9086 & 0.5578 & 0.6772 & 0.6343 & 0.6906 \\
    & MS-PS       & 0.8582 & 0.6269 & 0.9086 & 0.6698 & 0.7799 & 0.6623 & 0.7476 \\
\midrule
\multirow{2}{*}{Gemini-1.5}     
    & Single-Prompt & 0.6754 & 0.6287 & 0.7780 & 0.5784 & 0.7463 & 0.6119 & 0.6852 \\
    & MS-PS       & 0.8582 & 0.6213 & 0.9086 & 0.6698 & 0.7817 & 0.6996 & 0.7543 \\
\midrule
\multirow{2}{*}{Gemini-2.0}     
    & Single-Prompt & 0.4795 & 0.5634 & 0.8694 & 0.5840 & 0.7780 & 0.6922 & 0.7178 \\
    & MS-PS       & 0.8563 & 0.6269 & 0.9086 & 0.6716 & 0.7985 & 0.6679 & 0.7530 \\
\bottomrule
\end{tabular}
}
\caption{Comparison between Single-Prompt and MS-PS approaches across 4 LLMs. We report per-strategy accuracy (abbreviations defined in Figure \ref{fig:persuasion_strategies_tax}) and overall Micro F1 score.}
\label{tab:msps_vs_single_prompt}
\end{table*}

\subsection{Results}
The results in Table~\ref{tab:msps_vs_single_prompt} show a clear advantage of MS-PS over the single-prompt baseline across all four LLMs. For every model, MS-PS achieves higher accuracy on nearly all strategies, with particularly strong improvements for \textit{Distraction}, \textit{Simplification}, and \textit{Attack on reputation}, where single-prompt performance was relatively weaker.
In terms of overall performance, the Micro F1 score consistently increases when using MS-PS, confirming that the two-step approach more accurately detects persuasive strategies than the simpler single-prompt method. Since all metrics are computed against gold-standard human annotations, the superior performance of MS-PS can be interpreted as a validation for the quality of the strategy scores it generates. In other words, the experiment confirms that MS-PS produces scores that align closely with human judgment, supporting the use of these scores for downstream analysis of persuasive content.

\section{Extension to Other Datasets}
\label{appendix:extension_new_domains}

To further validate the generality of our strategy-aware framework, we extended MS-PS-MLP to two additional persuasion datasets: \textit{\textbf{Anthropic}} (described in Section \ref{sec:anthropic_dataset}) and \textit{\textbf{Persuasion for Good}} (described in Section \ref{sec:pfg_dataset}). Both datasets differ substantially from the \textit{Winning Arguments} setup, requiring regression instead of classification. Below, we describe the experimental setup and results in detail. We restricted this evaluation to \texttt{Gemma-3-12B} and \texttt{OpenAI-o3}, as these were the two best performing models in our main setup.  

\subsection{Anthropic}
For each sample, we prompted the LLM with the claim and argument and asked it to generate six persuasion-strategy scores. We used the same prompts from MS-PS for step 1 (see Appendix~\ref{appendix:strategy_analysis_prompt}) and a slightly modified prompt, tailored to this task, for step 2 (see Appendix~\ref{appendix:strategy_prompts_anthropic}). These scores, along with their mean, variance, entropy, and the initial pre-argument rating, formed the input to our regression model. The target was the post-argument rating.

The model architecture was identical to MS-PS-MLP used in classification, except that the output layer consisted of a single neuron (instead of two) to produce a continuous value, and the input dimension was reduced to accept only 6 scores instead of the two series of 6 scores used for binary classification.

We compared against two baselines (full prompts can be found in Appendix~\ref{appendix:strategy_prompts_anthropic}):
\begin{itemize}
    \item \textbf{Baseline-1}: a single-prompt LLM prediction of the post-argument rating.
    \item \textbf{Baseline-2}: a refined version of Baseline-1, explicitly instructing the model that arguments may decrease as well as increase agreement.
\end{itemize}

Results are reported in Table~\ref{tab:anthropic_results}. Accuracy (\texttt{Acc.}) is calculated by rounding the predicted values to the nearest integer, and then assessing if it was the same number as the gold label. MS-PS-MLP achieved significantly lower errors and higher $R^2$, demonstrating the value of strategy-aware reasoning.

\begin{table}[H]
    \centering
    \scriptsize
    \setlength{\tabcolsep}{4pt}
    \renewcommand{\arraystretch}{1.1}
    \begin{tabular}{lccccc}
        \toprule
        \textbf{Method} & \textbf{MSE($\downarrow$)} & \textbf{RMSE($\downarrow$)} & \textbf{MAE($\downarrow$)} & \textbf{$R^2$($\uparrow$)} & \textbf{Acc.($\uparrow$)} \\
        \midrule
        Baseline-1-Gemma & 4.3553 & 2.0869 & 1.6988 & -0.229 & 0.184 \\
        Baseline-2-Gemma & 3.5854 & 1.8935 & 1.5347 & -0.011 & 0.201 \\
        Baseline-1-o3 & 1.0068 & 1.0034 & 0.7394 & 0.716 & 0.374 \\
        Baseline-2-o3 & 0.9882 & 0.9941 & 0.7310 & 0.721 & 0.379 \\
        \textbf{MS-PS-MLP} & \textbf{0.6700} & \textbf{0.8185} & \textbf{0.6339} & \textbf{0.811} & \textbf{0.462} \\
        \bottomrule
    \end{tabular}
    \caption{Results on the Anthropic dataset for MS-PS-MLP compared to two baselines.}
    \label{tab:anthropic_results}
\end{table}

It is important to note that the Anthropic dataset was released in September 2024. As shown in Appendix~\ref{appendix:llms}, the knowledge-cutoff dates for Gemma-3-12B and OpenAI-o3 precede this release, so these models could not have been trained on Anthropic. Consequently, the strong MS-PS results on Anthropic cannot be attributed to pretraining leakage and therefore help alleviate concerns about dataset contamination.

\subsection{Persuasion for Good}
To avoid leakage, we excluded all persuadee turns and prompted the LLM only with the persuader turns. This prevents the model from simply reading the donation amount directly from the persuadee’s response. We then prompted an LLM to produce six persuasion-strategy scores per dialogue. We used the same prompts from MS-PS for step 1 (see Appendix~\ref{appendix:strategy_analysis_prompt}) and a slightly modified prompt, tailored to this task, for step 2 (see Appendix~\ref{appendix:strategy_prompts_pfg}). These scores, along with their mean, variance, entropy, and the initial pre-argument rating, formed the input to our regression model. The target label was the donation amount.  

The model architecture was identical to MS-PS-MLP used in classification, except that the output layer consisted of a single neuron (instead of two) to produce a continuous value, and the input dimension was reduced to accept only 6 scores instead of the two series of 6 scores used for binary classification.

We compared against two baselines (full prompts can be found in Appendix~\ref{appendix:strategy_prompts_pfg}):
\begin{itemize}
    \item \textbf{Baseline-1}: a single-prompt prediction of the donation amount.
    \item \textbf{Baseline-2}: a refined prompt discouraging overestimation of donations.
\end{itemize}

Results are reported in Table~\ref{tab:pfg_results}. Again, MS-PS-MLP strongly outperformed baselines, with orders of magnitude reduction in MSE.

\begin{table}[H]
    \centering
    \scriptsize
    \setlength{\tabcolsep}{4pt}
    \renewcommand{\arraystretch}{1.1}
    \begin{tabular}{lcccc}
        \toprule
        \textbf{Method} & \textbf{MSE($\downarrow$)} & \textbf{RMSE($\downarrow$)} & \textbf{MAE($\downarrow$)} & \textbf{$R^2$($\uparrow$)} \\
        \midrule
        Baseline-1-Gemma & 24352.63 & 156.05 & 68.43 & -1354.91 \\
        Baseline-2-Gemma & 11365.33 & 106.61 & 18.47 & -631.80 \\
        Baseline-1-o3 & 21978.06 & 148.25 & 61.02 & -993.21 \\
        Baseline-2-o3 & 9512.10 & 97.53 & 16.18 & -485.14 \\
        \textbf{MS-PS-MLP} & \textbf{21.41} & \textbf{4.63} & \textbf{2.48} & \textbf{-0.19} \\
        \bottomrule
    \end{tabular}
    \caption{Results on the Persuasion for Good dataset for MS-PS-MLP compared to two baselines.}
    \label{tab:pfg_results}
\end{table}

\begin{table*}[h]
\centering
\small
\resizebox{\textwidth}{!}{%
\begin{tabular}{|l|l|l|l|l|}
\hline
\textbf{API Model Name} & \textbf{Knowledge Cutoff Date} & \textbf{Access Details} & \textbf{License} & \textbf{Model Size} \\
\hline
meta-llama/Meta-Llama-3.1-8B-Instruct & Dec 2023 & DeepInfra API & Meta Llama 3 Community & 8B \\
google/gemma-3-12b-it & Aug 2024 & Hugging Face Transformers & Gemma Terms of Use & 12B \\
gemini-1.5-flash-002 & May 2024 & Google API & Commercial & Not Disclosed \\
gemini-2.0-flash & Aug 2024 & Google API & Commercial & Not Disclosed \\
o3-2025-04-16 & Jun 2024 & OpenAI API & Commercial & Not Disclosed \\
\hline
\end{tabular}%
}
\caption{Large Language Models used in our experiments.}
\label{tab:llms}
\end{table*}

\subsection{Prompts for Anthropic Experiments}
\label{appendix:strategy_prompts_anthropic}

We used the following prompts for the baselines and the MS-PS scoring procedure.  
After running Baseline-1, we observed that the models consistently predicted final ratings that were higher than the initial rating, regardless of the persuasiveness of the argument. To address this issue, in Baseline-2 we explicitly added the instruction that the new rating can be both higher or lower than the initial one. This modification resolved the problem and led to improved results.

\paragraph{Baseline-1}
\begin{quote}
\small
Given an original claim, an argument that tries to make the reader agree with the claim, and the initial rating given by the reader, you have to respond with a number from 1 to 7 (1 means strongly disagree, 7 strongly agree) based on the new rating that you think the reader gave after reading the argument. The goal is thus to understand how much you think that the claim succesfully convinced the reader to agree with the claim. Answer with only a number from 1 to 7, nothing else.
\end{quote}

\paragraph{Baseline-2}
\begin{quote}
\small
Given an original claim, an argument that tries to make the reader agree with the claim, and the initial rating given by the reader, you have to respond with a number from 1 to 7 (1 means strongly disagree, 7 strongly agree) based on the new rating that you think the reader gave after reading the argument. The goal is thus to understand how much you think that the claim succesfully convinced the reader to agree with the claim. Consider that the new rating can be both higher or lower than the initial one. Answer with only a number from 1 to 7, nothing else.
\end{quote}

\paragraph{MS-PS Strategy Scoring}
\begin{quote}
\small
Given an original claim and an argument that tries to make the reader agree with the claim, as well as an analyses on persuasion strategies used in the argument, you have to respond with a number from 1 to 10 based on the degree of persuasiveness of the argument, preceeded by a brief explanation on why you gave that score. Give your answer in the form of dictionary: \{"explanation": "Your answer. Brief explanation on the reasoning that you have followed.", "response": "Your answer. Give a score from 1 to 10."\}
\end{quote}

\subsection{Prompts for Persuasion for Good Experiments}
\label{appendix:strategy_prompts_pfg}

We used the following prompts for the baselines and the MS-PS scoring procedure.  
During initial experiments, we observed that the models often refused to answer, claiming that it was against their ethical guidelines to process dialogues where a persuader tries to convince someone to donate money. To mitigate this issue, we clarified in the prompts that the dataset was created for the good intention of donating for a charity. This addition reduced refusals and allowed the models to proceed with the task.  
Later, we noticed that explicitly providing the minimum and maximum possible values of donations (0 to 700, as found in the dataset) biased the models toward producing overly high predictions. To address this, we removed that sentence in Baseline-2, which improved the results.

\paragraph{Baseline-1}
\begin{quote}
\small
Given a text taken from the dataset Persuasion For Good (created for the good intention of donating for a charity) where a persuader tries to convince a user to donate for a charity (user turns are hidden), you have to respond with a number indicating the dollars you think the persuadee donated. It can be a number from 0 to 700, but consider that usually people donate around 2 dollars. Answer with only the number of dollars you think were donated, nothing else.
\end{quote}

\paragraph{Baseline-2}
\begin{quote}
\small
Given a text taken from the dataset Persuasion For Good (created for the good intention of donating for a charity) where a persuader tries to convince a user to donate for a charity (user turns are hidden), you have to respond with a number indicating the dollars you think the persuadee donated. It can whatever number, but consider that usually people donate around 2 dollars. Answer with only the number of dollars you think were donated, nothing else.
\end{quote}

\paragraph{MS-PS Strategy Scoring}
\begin{quote}
\small
Given a text taken from the dataset Persuasion For Good (created for the good intention of donating for a charity) where a persuader tries to convince a user to donate for a charity (user turns are hidden), as well as an analyses on persuasion strategies used in the text, you have to respond with a number from 1 to 10 based on the degree of persuasiveness of the text, preceeded by a brief explanation on why you gave that score. Give your answer in the form of dictionary: \{"explanation": "Your answer. Brief explanation on the reasoning that you have followed.", "response": "Your answer. Give a score from 1 to 10."\}
\end{quote}

\section{Strategy Impact Across Topics}
\label{appendix:strategy_impact}

While the main focus of this work is on improving predictive performance through strategy-guided reasoning, the structured outputs produced by \textsc{MS-PS} also enable a more fine-grained analysis of how different persuasion strategies relate to persuasiveness across content domains. In this appendix, we leverage this structure to investigate the relative importance of individual strategies within and across topics.

\subsection*{Analysis Setup}

For each model, topic, and persuasion strategy, we compute the difference between the average strategy score assigned to successful messages and that assigned to unsuccessful messages. Formally, for a given strategy \( s \), we define:

\[
\Delta s = \mathbb{E}[s_{\text{success}}] - \mathbb{E}[s_{\text{failure}}],
\]

where \( s_{\text{success}} \) and \( s_{\text{failure}} \) denote the scores assigned by \textsc{MS-PCoT} to messages that do and do not succeed in changing the original poster’s view, respectively.

This quantity provides a quantitative estimate of how strongly the presence of a given strategy correlates with persuasiveness in a particular domain. Higher values of \( \Delta s \) indicate that the model consistently attributes greater strategic strength to winning messages, suggesting that the corresponding strategy may be particularly salient or effective in that context. Conversely, values close to zero (or negative) indicate limited or inconsistent association with success.

\subsection*{Results}

Table~\ref{tab:strategy_impact_topics} reports \( \Delta s \) values for all models, aggregated both across topics and per topic. Several trends emerge. First, the magnitude and ranking of strategies vary substantially across domains, indicating that persuasiveness is not driven by a single universal strategy. Second, we note that different strategies are more or less effective on the different topics: for exemple, \emph{Attack on Reputation} and \emph{Simplification} are particularly effective for Topic 0, and \emph{Justification} and \emph{Distraction} for Topic 1. 
Finally, different models emphasize different strategic cues, suggesting model-specific biases in how persuasion signals are perceived.

\begin{table*}[t]
    \centering
    \scriptsize
    \setlength{\tabcolsep}{3pt}
    \renewcommand{\arraystretch}{1.05}
    \begin{tabular}{llcccccc}
        \toprule
        \textbf{Model} & \textbf{Topic} & \textbf{AR} & \textbf{J} & \textbf{S} & \textbf{D} & \textbf{C} & \textbf{MW} \\
        \midrule
        Llama & All & 0.226 & 0.231 & 0.283 & 0.116 & 0.309 & 0.304 \\
        Llama & 0   & 0.289 & 0.277 & 0.235 & 0.163 & 0.386 & 0.319 \\
        Llama & 1   & 0.546 & 0.397 & 0.609 & 0.425 & 0.575 & 0.592 \\
        Llama & 2   & 0.051 & 0.244 & 0.201 & -0.004 & 0.111 & 0.115 \\
        Llama & 3   & 0.116 & 0.060 & 0.155 & -0.030 & 0.255 & 0.268 \\
        \midrule
        Gemma & All & 0.436 & 0.521 & 0.457 & 0.392 & 0.217 & 0.357 \\
        Gemma & 0   & \textbf{0.627} & 0.467 & 0.564 & \textbf{0.709} & 0.464 & 0.564 \\
        Gemma & 1   & 0.494 & \textbf{0.793} & 0.626 & 0.563 & 0.270 & 0.374 \\
        Gemma & 2   & 0.321 & 0.440 & 0.385 & 0.154 & 0.030 & 0.248 \\
        Gemma & 3   & 0.373 & 0.438 & 0.326 & 0.279 & 0.189 & 0.313 \\
        \midrule
        Gemini & All & 0.466 & 0.530 & 0.416 & 0.486 & 0.431 & 0.395 \\
        Gemini & 0   & 0.530 & 0.416 & 0.410 & 0.584 & 0.506 & 0.313 \\
        Gemini & 1   & 0.506 & 0.695 & 0.322 & 0.460 & \textbf{0.586} & 0.621 \\
        Gemini & 2   & 0.278 & 0.444 & 0.521 & 0.440 & 0.321 & 0.359 \\
        Gemini & 3   & 0.579 & 0.575 & 0.386 & 0.481 & 0.373 & 0.322 \\
        \midrule
        Gemini-2 & All & 0.299 & 0.309 & 0.316 & 0.321 & 0.338 & 0.278 \\
        Gemini-2 & 0   & 0.398 & 0.388 & 0.364 & 0.303 & 0.521 & 0.380 \\
        Gemini-2 & 1   & 0.293 & 0.437 & 0.385 & 0.397 & 0.351 & 0.305 \\
        Gemini-2 & 2   & 0.286 & 0.261 & 0.261 & 0.248 & 0.248 & 0.107 \\
        Gemini-2 & 3   & 0.245 & 0.206 & 0.288 & 0.352 & 0.288 & 0.356 \\
        \midrule
        o3 & All & 0.499 & 0.475 & 0.457 & 0.503 & 0.455 & 0.513 \\
        o3 & 0   & 0.392 & 0.422 & 0.464 & 0.476 & 0.416 & 0.428 \\
        o3 & 1   & 0.489 & 0.575 & \textbf{0.644} & 0.586 & 0.558 & \textbf{0.661} \\
        o3 & 2   & 0.573 & 0.427 & 0.397 & 0.483 & 0.389 & 0.487 \\
        o3 & 3   & 0.511 & 0.485 & 0.373 & 0.481 & 0.472 & 0.489 \\
        \bottomrule
    \end{tabular}
    \caption{Difference \( \Delta s \) between average strategy scores for successful and unsuccessful messages. AR = Attack on Reputation, J = Justification, S = Simplification, D = Distraction, C = Call, MW = Manipulative Wording.}
    \label{tab:strategy_impact_topics}
\end{table*}

\section{LLMs Used in Experiments}
\label{appendix:llms}

In our experiments, we used a diverse set of Large Language Models to ensure broad applicability and test the robustness of our method across different architectures, sizes, and access modalities. We included both commercial API-based models and open-weight models, trying to balance accessibility and performance.
In particular, we evaluated the system using Meta-Llama-3.1 (8B Instruct), Gemma-3-12B, Gemini 1.5 Flash-02, Gemini 2 Flash and OpenAI-o3.
Table \ref{tab:llms} provides details on each model’s access, license, size, and knowledge cutoff. 

\end{document}